%% file: main.tex
\documentclass[lettersize,journal]{IEEEtran}
\usepackage[utf8]{inputenc}
\usepackage[margin=1in]{geometry}
\usepackage{algorithm}
\usepackage{algorithmic}
\usepackage{amssymb}		
\usepackage{graphicx}		
\usepackage{amsmath}
\usepackage{comment}
\usepackage{hhline}
\usepackage{array}

\usepackage{booktabs}
\usepackage{multirow}
\usepackage{graphicx}
\usepackage{bm}
\usepackage{subfig}
\usepackage{tablefootnote}

\usepackage[normalem]{ulem}
\usepackage{color}
\usepackage{soul}

\usepackage[font=small,labelfont=bf]{caption}

\usepackage{hyperref}
\hypersetup{
  colorlinks=true, 
  urlcolor=blue,   
  linkcolor=blue,  
  citecolor=blue   
}

\usepackage{balance}

\providecommand{\figref}[1]{Fig.~\ref{#1}}
\providecommand{\tabref}[1]{Table~\ref{#1}}
\providecommand{\secref}[1]{\textsection\ref{#1}}

\usepackage{user_acronyms}

\usepackage{cite}
\usepackage[numbers]{natbib}

\newcommand{\change}[1]{{\color{black}#1}}
\newcommand{\changefeb}[1]{{\color{blue}#1}}

\title{A New Wave in Robotics: Survey on Recent mmWave Radar Applications in Robotics}
\author{Kyle Harlow$^1$, \IEEEmembership{Student Member,~IEEE}, Hyesu Jang$^2$, \IEEEmembership{Student Member,~IEEE}, Timothy D. Barfoot$^3$, \IEEEmembership{Fellow,~IEEE}, Ayoung Kim$^2$, \IEEEmembership{Member,~IEEE}, Christoffer Heckman$^1$ \IEEEmembership{Senior Member,~IEEE}%
\thanks{$^1$Autonomous Robotics and Perception Group at the
        University of Colorado Boulder, Boulder Colorado, USA. Corresponding author: 
        {\tt\small christoffer.heckman@colorado.edu}}
\thanks{$^2$ Robust Perception and Mobile Robotics Lab at Seoul National University, Seoul, S. Korea}
\thanks{$^3$ Autonomous Space Robotics Laboratory at the University of Toronto, Toronto, Canada}
}
\date{January 2021}

\begin{document}

\maketitle

\begin{abstract}

    We survey the current state of \ac{mmWave} radar applications in robotics with a focus on unique capabilities, and discuss future opportunities based on the state of the art. \ac{FMCW} mmWave radars operating in the 76--81GHz range are an appealing alternative to lidars, cameras and other sensors operating in the near visual spectrum.  Radar has been made more widely available in new packaging classes, more convenient for robotics and its longer wavelengths have the ability to bypass visual clutter such as fog, dust, and smoke.  
    We begin by covering radar principles as they relate to robotics. We then review the relevant new research across a broad spectrum of robotics applications beginning with motion estimation, localization, and mapping. We then cover object detection and classification, and then close with an analysis of current datasets and calibration techniques that provide entry points into radar research. 
    
\end{abstract}

\section{Motivation and Background}
\input{intro.tex}
\section{Radar Preliminaries}
\label{sec:radar-preliminaries}
\input{radar_basics_simplified.tex}

\section{Motion Estimation}
\label{sec:motion-estimation}
\input{motion_estimation.tex}

\section{Localization and Mapping}
\label{sec:localization}
\input{relocalization_and_mapping.tex}

\section{Object Classification}
\label{sec:object-detection}
\input{object_classification.tex}

\section{Datasets, Simulation, and Calibration}
\label{sec:datasets}
\input{datasets_and_calibration}



\section{Open Problems}
\input{openproblems.tex}

\section{Conclusion}

\input{conclusion.tex}

\section{Acknowledgements}
This work was supported through the DARPA Subterranean Challenge cooperative agreement HR0011-18-2-0043, the National Science Foundation \#1764092, and the NRF Korea (Grant No. RS-2023-00241758).

\small
\bibliographystyle{IEEEtranN}
\bibliography{IEEEabrv,refs}

\newcommand{\bioshot}[1]{\includegraphics[width=1in,height=1.25in,clip,keepaspectratio]{#1}}

\begin{IEEEbiography}
[\bioshot{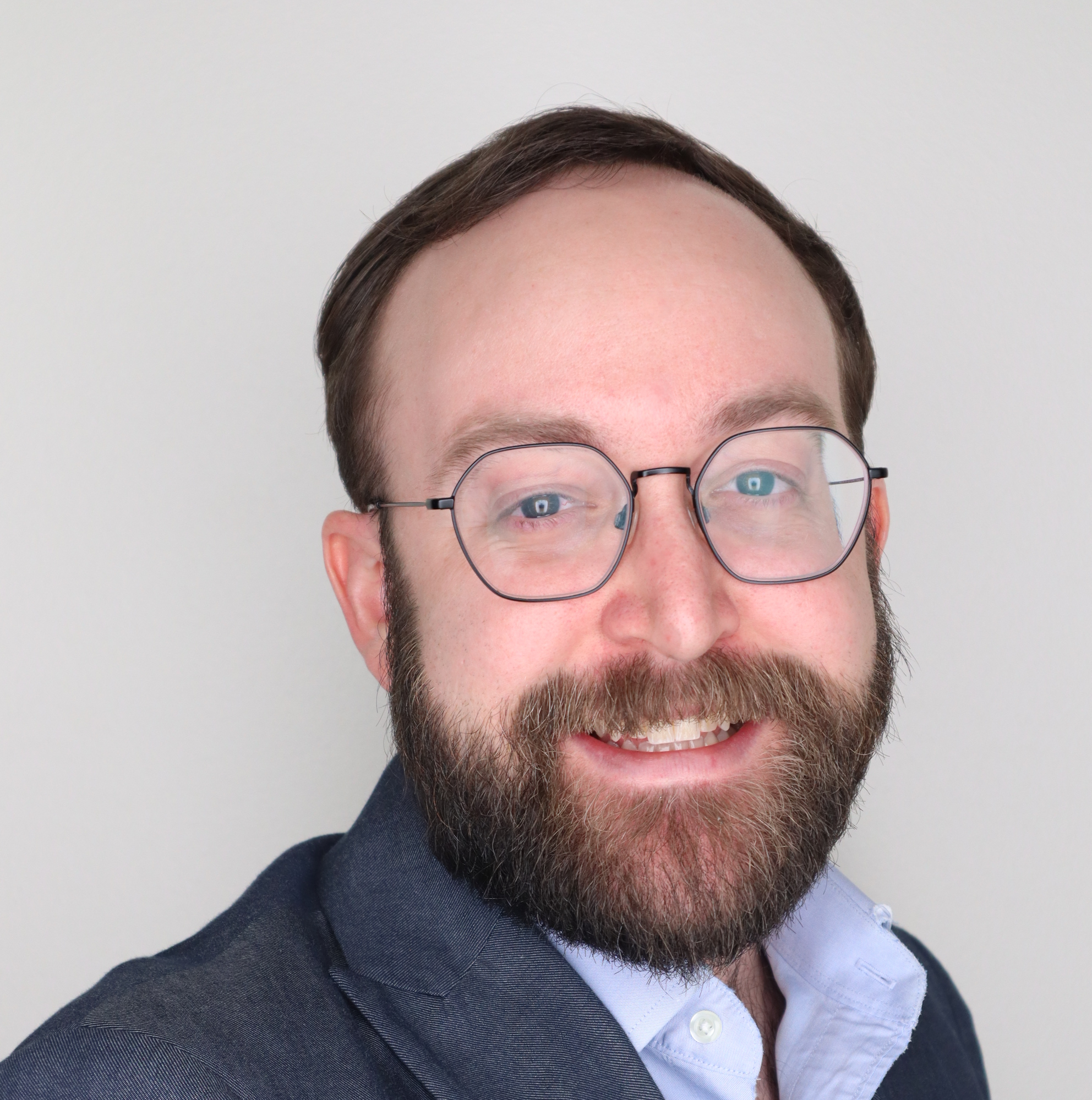}]{Kyle Harlow}(S'17--S'19--Su'24) Received a B.S. in Applied Mathematics, a B.S./M.S. in Electrical Engineering, and a Ph.D. in Computer Science from the University of Colorado -- Boulder, Boulder, CO, USA. At Boulder, he worked on the DARPA Subterranean Challenge where his team, Team MARBLE, placed 3rd overall. His research is in developing mapping and odometry solutions using dense millimeter wave radar.
\end{IEEEbiography}%

\begin{IEEEbiography}[\bioshot{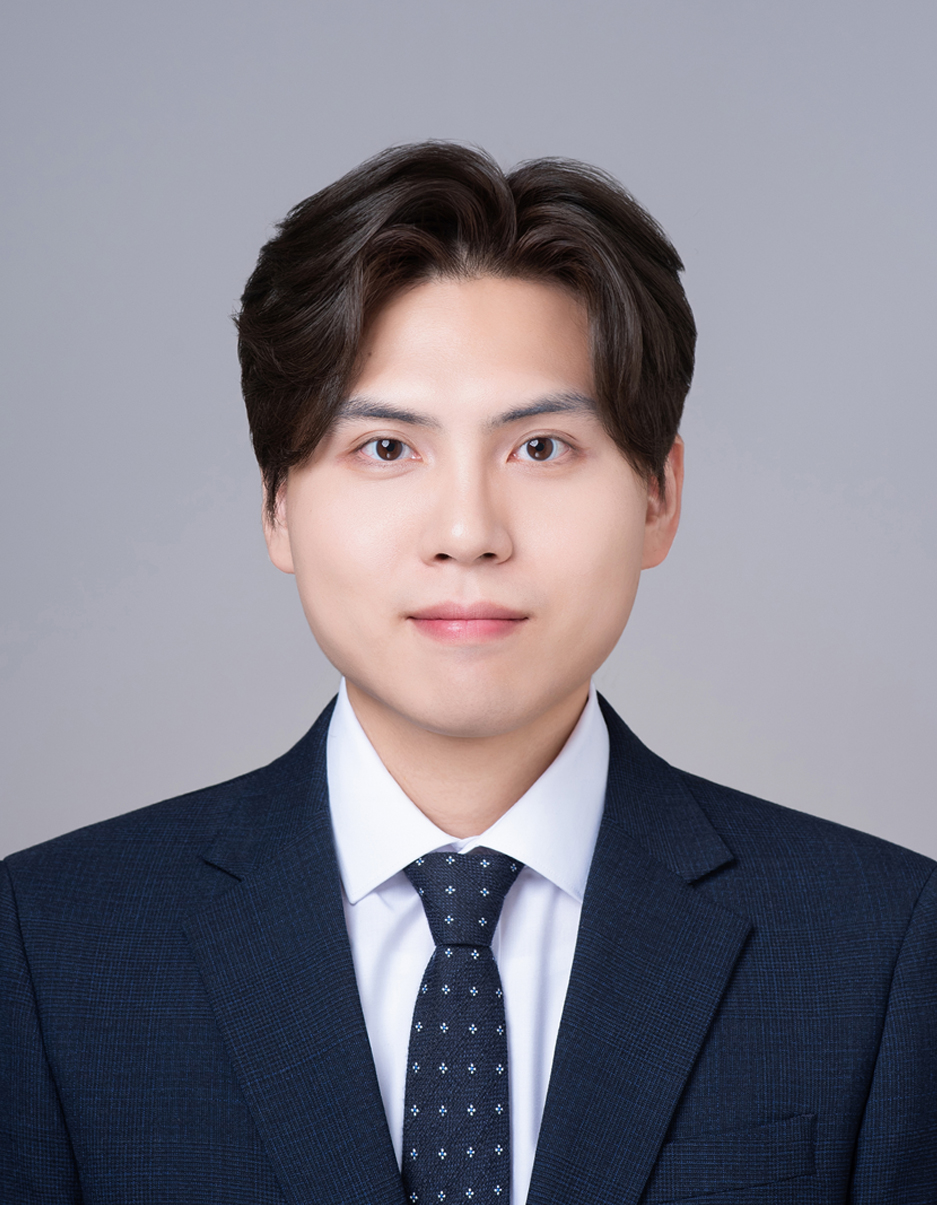}]{Hyesu Jang}%
(S'19) received the B.S. and M.S. degrees in civil and environmental engineering from Korea Advanced Institute of Science and Technology (KAIST) in 2018 and 2020, and the Ph.D. degree in mechanical engineering from Seoul National University (SNU) in 2024. Currently, he is a Post doctoral researcher in the Institute of Advanced Machines and Design, Seoul National University (SNU). His research interests include radar SLAM and maritime mapping.
\end{IEEEbiography}%

\begin{IEEEbiography}
[\bioshot{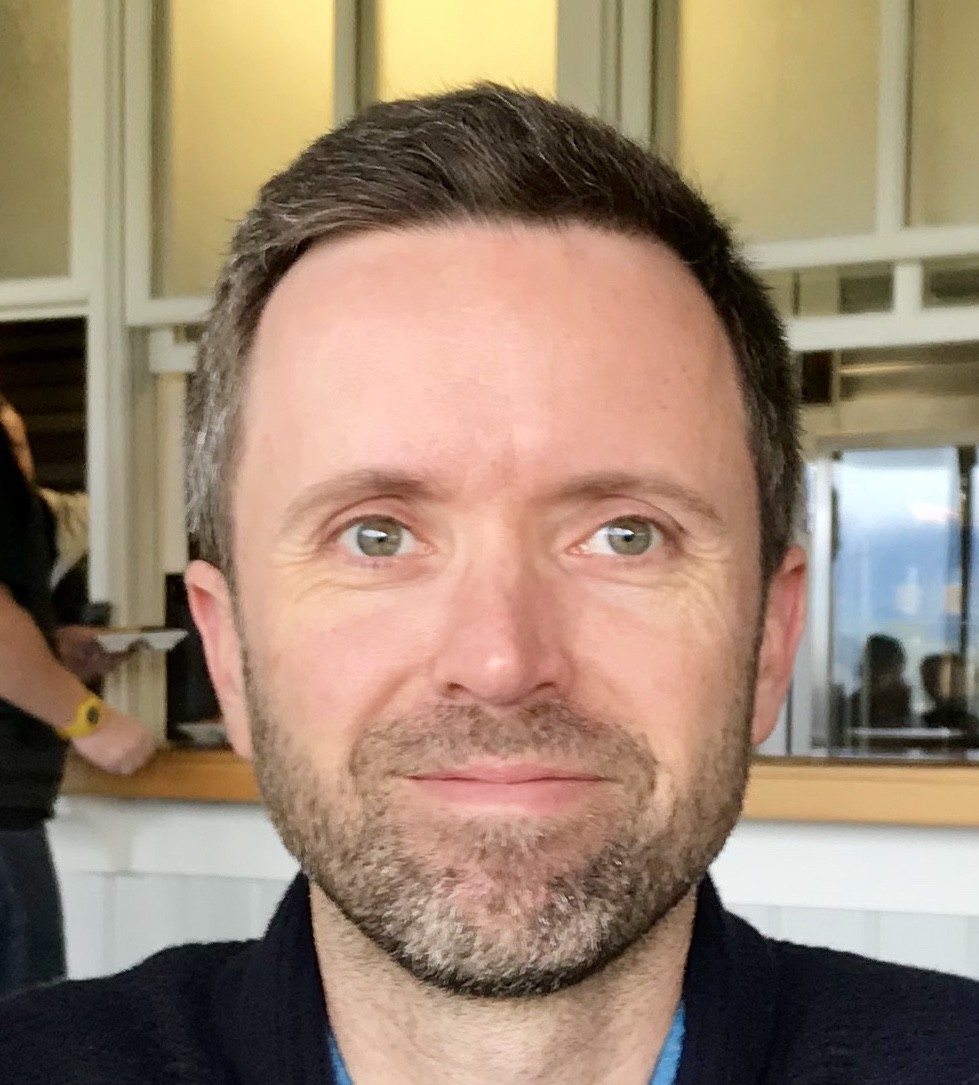}]{Timothy D. Barfoot} received the B.A.Sc. degree in engineering science from University of Toronto, Toronto, ON, Canada, in 1997 and the Ph.D. degree in aerospace science and engineering from University of Toronto, in 2002. He is a Professor with the University of Toronto Robotics Institute, Toronto, ON, Canada. He works in the areas of guidance, navigation, and control of autonomous systems in long-term deployments and unstructured environments.
\end{IEEEbiography}%

\begin{IEEEbiography}[\bioshot{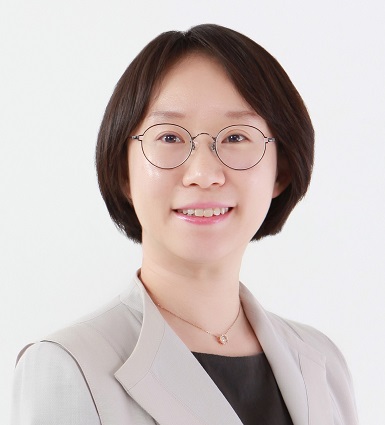}]{Ayoung Kim}%
(S'08--M'13--S'23) received the B.S. and M.S. degrees from Seoul National University (SNU) in 2005 and 2007, and the Ph.D. degree from the University of Michigan (UM), Ann Arbor 2012. She was an associate professor at Korea Advanced Institute of Science and Technology (KAIST) from 2014 to 2021. Currently, she is an associate professor at SNU.
\end{IEEEbiography}%

\begin{IEEEbiography}
[\bioshot{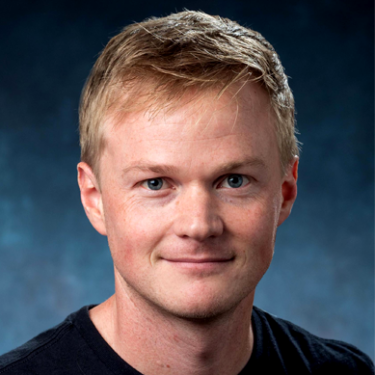}]{Christoffer R.\ Heckman} received the B.S. degree in mechanical engineering from University of California Berkeley in 2008 and the Ph.D.\ degree in theoretical and applied mechanics from Cornell University in 2012. He is an Associate Professor with the University of Colorado Boulder. He works in the area of autonomous field robotics and machine learning techniques in sensor fusion.
\end{IEEEbiography}%


\end{document}

%% file: intro.tex
\textbf{Ra}dio \textbf{d}etection \textbf{a}nd \textbf{r}anging or \textit{radar}, the process of transmitting and receiving bespoke electromagnetic pulses to determine the range and bearing of objects of interest, has a rich history in engineering solutions. Despite their solid application in measuring weather, tracking targets, mapping planets, and constituting safety systems for automotives~\citep{schneider2005automotive}, radars as perceptual sensors in robotics have been relatively overlooked compared to other sensors with shorter wavelengths (e.g., cameras). 


Traditionally, robotics systems have relied on light-based sensors, such as cameras and lidar, for building a representation of their environments. 
Through these sensors, advanced techniques for localization and identification have been developed \cite{orb-tro15,engel2017direct,loam-rss14,liosam-iros20}.
Unfortunately, cameras and lidars are not universally functional when encountering illuminationally or structurally degenerate cases. Despite successive efforts on photometric calibration \cite{engel2017direct}, image enhancement \cite{dark-ral20}, and camera exposure control \cite{expctrl-tro20}, these two sensors in the near-visible spectrum suffer severe degradation in a harsh environment. 

\begin{figure}[t!]
  \centering
  \includegraphics[width=\linewidth]{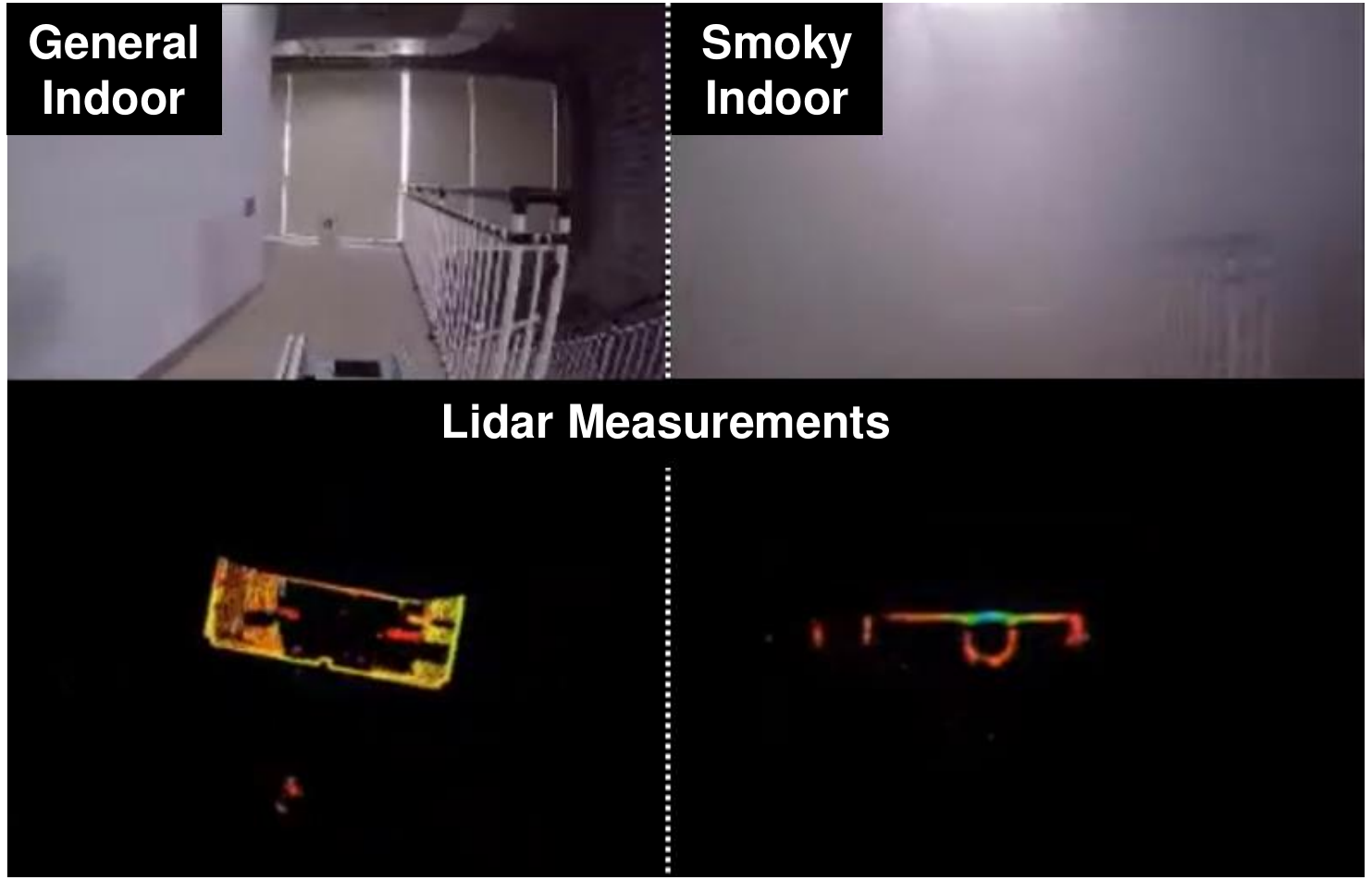}
  \caption{\ac{VDEs} cause significant degradation of both lidar and camera based measurements in autonomous robotics. Camera and lidar measurements in a smoke-filled indoor environment eliminate most identifiable structure and features required for navigation or 
  classification.}
  \label{fig:dust}
\end{figure}

\input{tab_sensor_cmp.tex}

As in sample measurements in \figref{fig:dust}, challenging navigation conditions such as smoke, dust, fog, rain, and snow critically deteriorate the camera \cite{kramer2020mav_fog} and lidar \cite{fire-13} but not the radar. Using longer wavelengths, radars theoretically penetrate through the varying small particulate matter.
We present a qualitative and general comparison of various exteroceptive sensors in \tabref{tab:sensor_basics} to address the specific strengths and weaknesses of each sensor.

Radar was first developed in the late 19th century and early 20th century. It demonstrated an ability to track large ships and airplanes and became a key factor in tracking targets during World War II \cite{detection152ranging}. During the development of tracking radar and beyond, the sensors were adapted to a variety of other purposes including weather monitoring \cite{whiton1998history}, and radio astronomy \cite{jansky1933electrical}. Applications of radar have become incredibly broad as technology and computing resources have enabled radar to be deployed and analyzed more easily. Radar has been used in space automation for many purposes from the Apollo program \cite{rozas1972apollo} to recent International Space Station operations \cite{mokuno2004orbit}.

\begin{figure*}[htb]
\centering
\begin{tabular}{cccccc}
    \includegraphics[height=2.7cm]{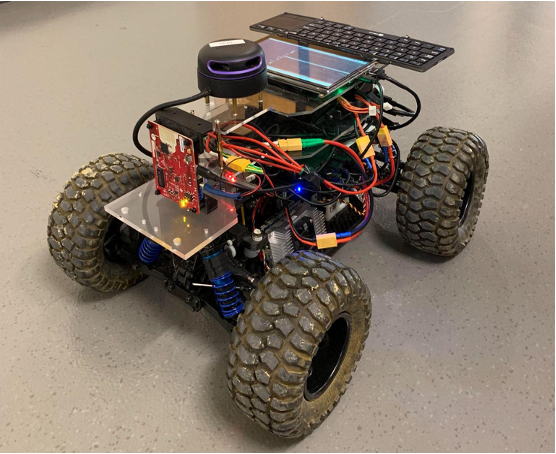} &
    \includegraphics[height=2.7cm]{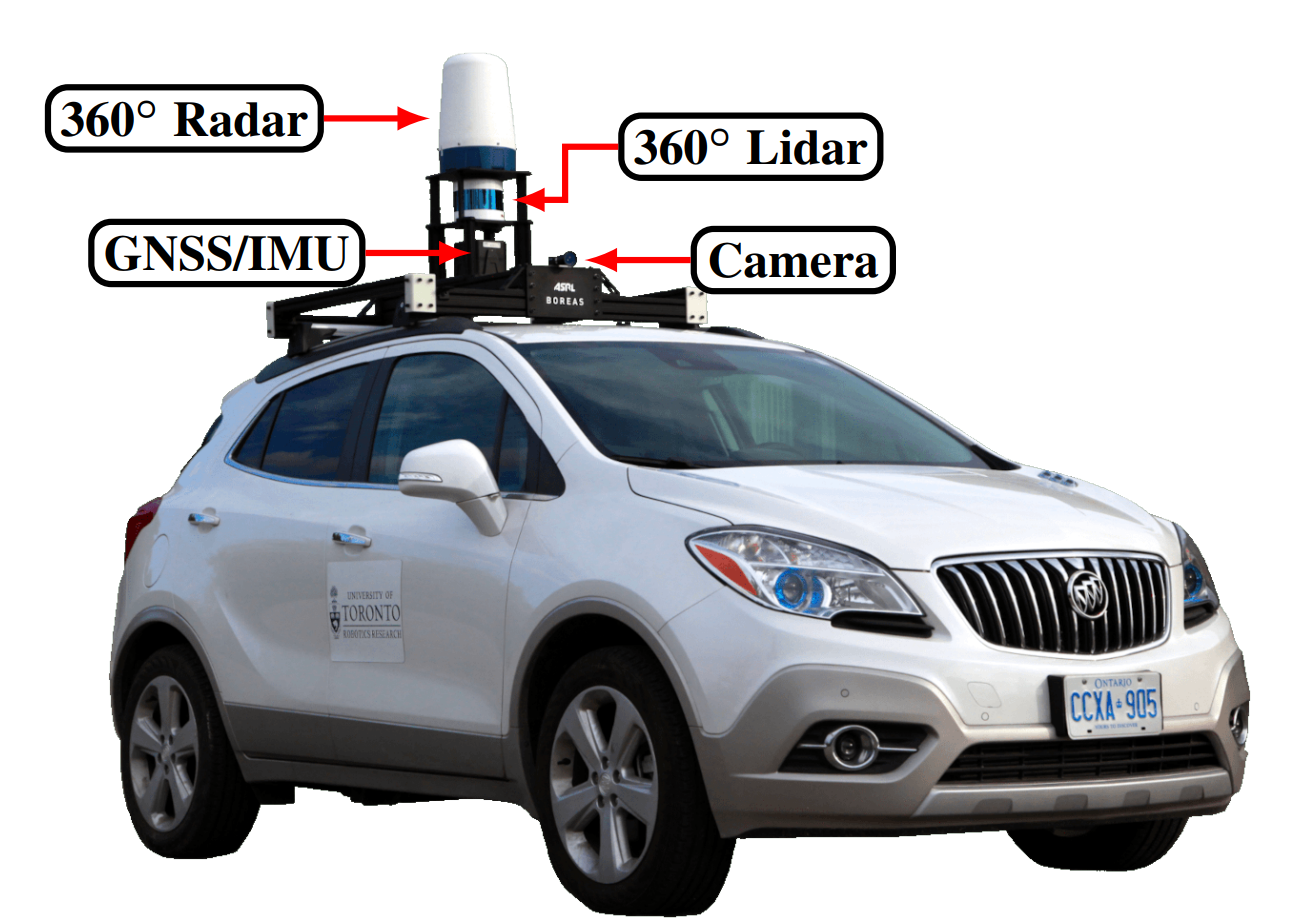} &
    \includegraphics[height=2.7cm]{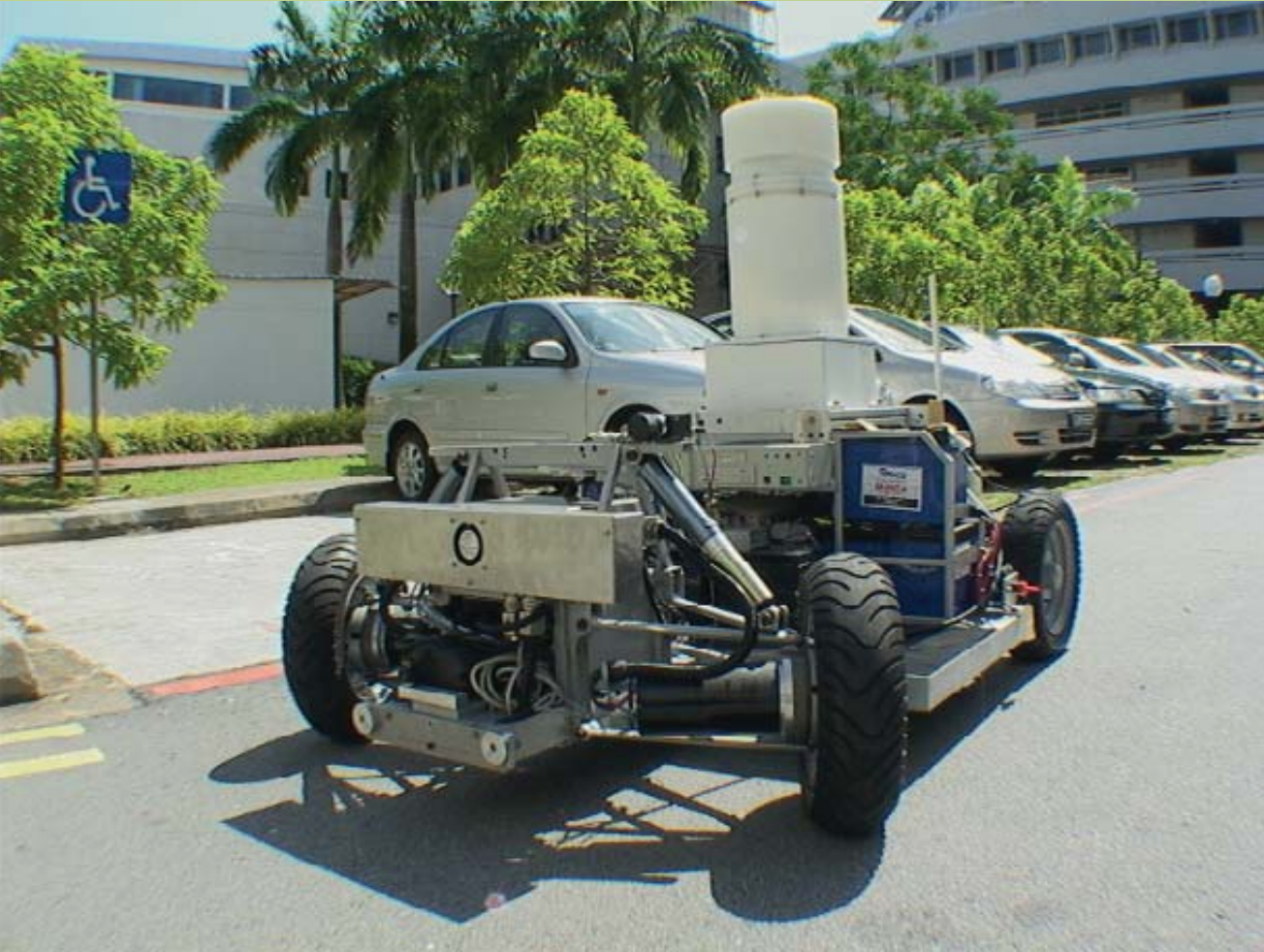} &
    \includegraphics[height=2.7cm]{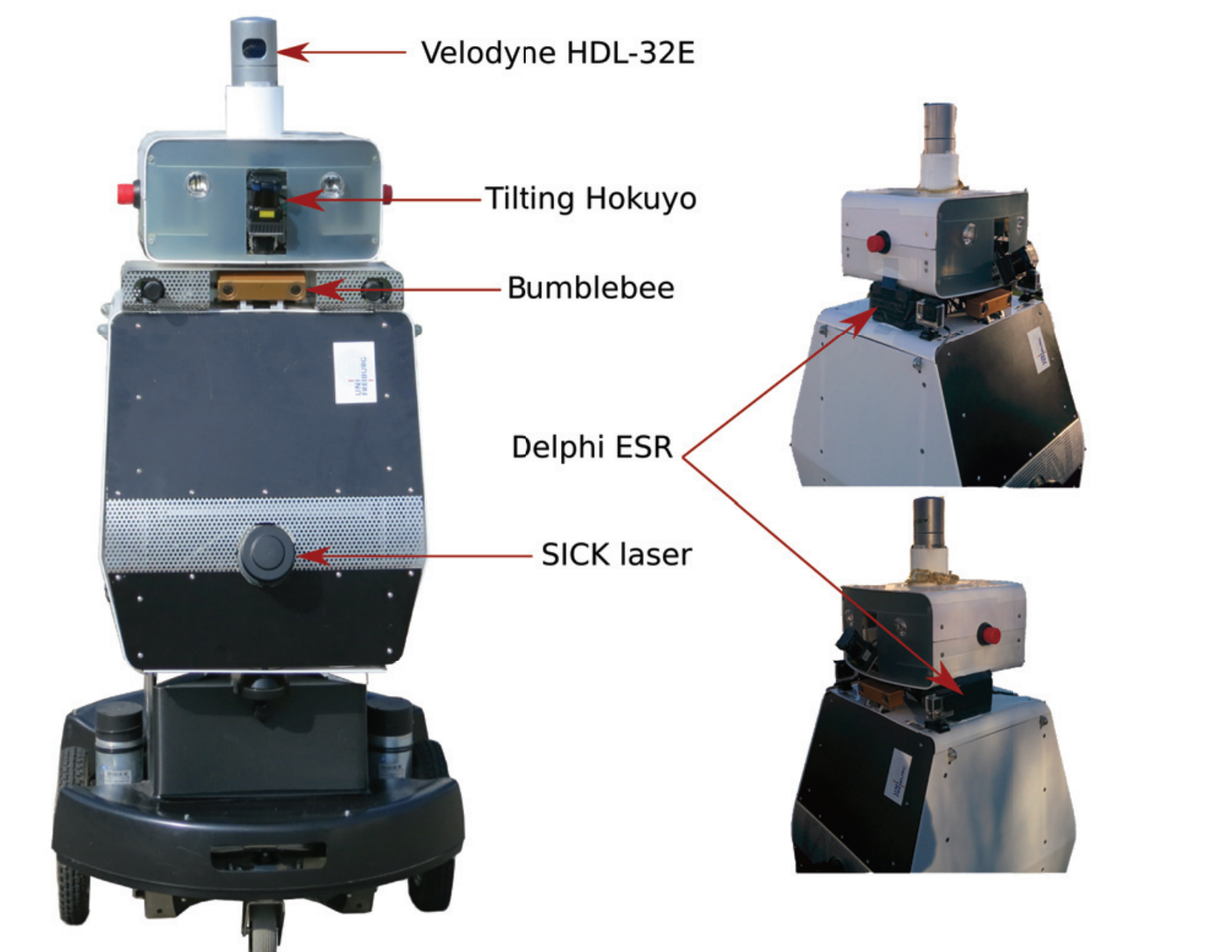} \\
\end{tabular}
\caption{Example robotic platforms with attached radar systems. From left to right: Fast moving Parkour-Car from the University of Colorado--Boulder equipped with a Texas Instruments system-on-chip radar. Boreas data collection vehicle from the University of Toronto equipped with a modern Navtech scanning radar \cite{burnett2022boreas}.  Ground vehicle from Nanyang Technical University using an older Navtech scanning radar \cite{adams2007autonomous}.  Street-crossing robot attached with two Delphi system-on-chip radar \cite{radwan2020multimodal}.}
\label{fig:radar_on_robots}
\end{figure*}

Given the robust sensing capabilities available, radar is an interesting sensor to fuse with existing information pipelines or to use as a standalone sensor for the various metric and semantic tasks robots must undertake for practical applications.
In this regard, some meaningful works in radar have been widely deployed, including autonomous vehicles \cite{venon2022millimeter}, \ac{DOT} \cite{pearce2023multi}, and \ac{SLAM} \cite{adams12}, while their usefulness beyond that has only begun to be explored. Furthermore, these radar sensors are now being packaged to meet automotive reliability requirements.
Radar has also been installed on mobile robot platforms in the desire to reduce the sensor payload in many applications, initiating an expanded role in localization, mapping, and object classification more broadly. Examples of several robots with attached radar are shown in \figref{fig:radar_on_robots}.  In parallel, recent efforts have aimed to support these new roles for the radar sensor, specifically with respect to algorithm development. As each new application is proposed, specific challenges provided by radar 
are addressed by innovative algorithms. 

This survey seeks to address mmWave radar applications to autonomous robotics. Interest in radar has skyrocketed recently as new packaging classes have made the sensor more available and convenient for various platforms. The sensor also provides data that makes it a viable alternative or complementary sensor to lidars or cameras (henceforth referred to as vision-based sensors due to the optical frequencies involved in their operation).

\textbf{\textit{Increased interest for radar in robotics: }} Since at least 2005, mmWave radar has been deployed in vehicles for assisted cruise control and automatic breaking systems \cite{schneider2005automotive}. This generated a demand for new packaging and cheaper versions of previously expensive equipment. 
Radar remained largely an automotive sensor for several years afterward; however, with the explosion of research in the late 2000s into autonomous vehicles, robotics perception researchers began utilizing these new devices in broader contexts. 
For instance, shifting from industrial application, spinning radar mountable to a mobile platform was developed by the \ac{ACFR} for 3D mapping and deployed in a mining application \cite{acfr-jfr07}. Martin Adams, while at Nanyang Technical University, produced several methods for localizing and grid-mapping with industrial radar mounted on ground vehicles surveyed in \cite{adams2007autonomous}.  
Meanwhile, \citet{de2017survey} provided a comparison of various ranging sensors for relative vehicle positioning that included radar, lidar, and vision based solutions. That, along with promising sparse localization techniques from the likes of \citet{schuster2016landmark,schuster2016robust,cen2018precise,cen2019radar}, demonstrated radar's potential viability as a competing robotics sensor and popularized it amongst some research groups. 


\textbf{\textit{Alternative and complementary sensor: }} Radar presents many unique challenges that we detail briefly in \secref{sec:radar-challenges}. Yet, as researchers have begun to address these challenges, they have also discovered the benefits provided by radar including longer ranges, object penetration, and direct velocity measurements. These will be explored thoroughly through the course of this survey. Since radar is not a commonplace sensor on robotic platforms today, we provide some mathematical preliminaries focusing on use cases, filtering methods, and datatypes. This differs from automotive-focused surveys such as \cite{de2017survey, venon2022millimeter}, where preliminaries instead focus more on radar physics and electrical properties. We later present algorithm-focused sections describing some early radar motion estimation work through feature-based and direct methods. We then provide an overview of state-of-the-art localization and mapping techniques utilizing radar. Next we provide an analysis of radar object detection and classification algorithms, separated by methods where radar is fused with  other sensors, or is used as a unique sensor. We then present the diverse datasets available containing radar data so 
readers may experiment independently, along with an analysis of radar calibration for 
aligning sensors. 

Throughout each section, this survey endeavors to provide both a detailed overview of the state-of-the-art in each category. Categories were chosen to provide starting points with sufficient detail but without overwhelming the reader with potential citations. We hope to show how radar can be used to supplement existing sensors such as cameras and lidar or act as a standalone sensor in a variety of metric and semantic robotics applications. 

%% file: tab_sensor_cmp.tex
\begin{table*}
\resizebox{\linewidth}{!}{
\begin{tabular}{l|llll|l|ll} 
\hline
\textbf{Sensor}                         & \multicolumn{4}{l|}{\textbf{Basic Sensor Information}}                                                                   & \multicolumn{1}{l|}{\textbf{Data Generated}}                              & \multicolumn{2}{l}{\textbf{Operation Capability}}      \\ 
\hline\hline
                                        & \textbf{Data Types}                                        & \textbf{Range}                           & \textbf{Data Density$^{\dagger}$} & \textbf{Resolution}                               & \textbf{Velocity} & \textbf{Darkness}     & \textbf{VDE}                    \\ 
\cline{1-8}
\textbf{Cameras}                        & 2D Images (RGB/BW)                                         & N/A                                      & Dense                 & 1mil-2mil px                                      & No                & \changefeb{Limited}   & \changefeb{Limited}             \\
\textbf{Flash Lidar/Depth Sensors}      & Depth Maps                                                 & 0.5m-5m                                  & Dense                 & 38k-50k px                                        & No                & Full                  & Occluded                        \\
\textbf{Lidar}                          & 3D Point Clouds                                            & 1m-200m                                  & Sparse                & 1mm,$0.17^{\circ}$Az.,,$0.35^{\circ}$El.           & No                & Full                  & Occluded                        \\
\textbf{2D Scanning Radar}              & Polar Range Map                                            & 50m/200m\textsuperscript{*}              & Dense                 & 0.06m/0.25m\textsuperscript{*},$1.8^{\circ}$Az.   & No                & Full                  & Full                            \\
\multirow{2}{*}{\textbf{3D SoC Radar }} & \begin{tabular}[c]{@{}l@{}}Spectral Heatmap\\\end{tabular} & 1m-50m/200m\textsuperscript{*}           & Dense                 & 0.04m/1m\textsuperscript{*},$1^{\circ}$Az.,$2^{\circ}$-$20^{\circ}$El.\textsuperscript{*}                       & Yes               & Full                  & Full                            \\
                                        & 3D Point Clouds                                            & 1m-50/200m\textsuperscript{*}            & Sparse                & 0.04m/1m\textsuperscript{*},$1^{\circ}$Az.,$5^{\circ}$-$20^{\circ}$El.\textsuperscript{*}                                            & Yes               & Full                  & Full                           
\end{tabular}}
\caption{Capabilities and basic sensor information of various exteroceptive sensors commonly used in robotics applications summarized in terms of optimal sensor ranges, data density, and data resolution. It additionally describes the dimensions each sensor can identify and how robust sensors are to visual degradation in their environment. $^{\dagger}$Dense refers to all values in a sensors field being populated, sparse means only some values within a sensors field are populated. \textsuperscript{*}Radar sensors have various configurations depending on antenna and waveform parameters, leading to differences in range and resolution.}\label{tab:sensor_basics}
\vspace{-7mm}
\end{table*}

%% file: radar_basics_simplified.tex
To further answer the question on why a reader should consider using mmWave radar we next discuss basic operations, antenna and datatypes, and challenges as they relate to robotics applications. We outline each in the following sections and designate how radar differentiates itself from other rangefinding sensors. 

\subsection{mmWave Radar}
\label{sec:mmwave-radar}

MmWave radar is designated by radar systems whose electromagnetic wavelengths are between 1 and 10mm with frequency ranging from 30 to 300GHz. Many governments have designated specific bandwidths around 76-81GHz for automotive applications such as ADAS systems and as such most commercial mmWave radar operate in this frequency. 


\begin{figure}[!t]
\centering
\includegraphics[width=\columnwidth]{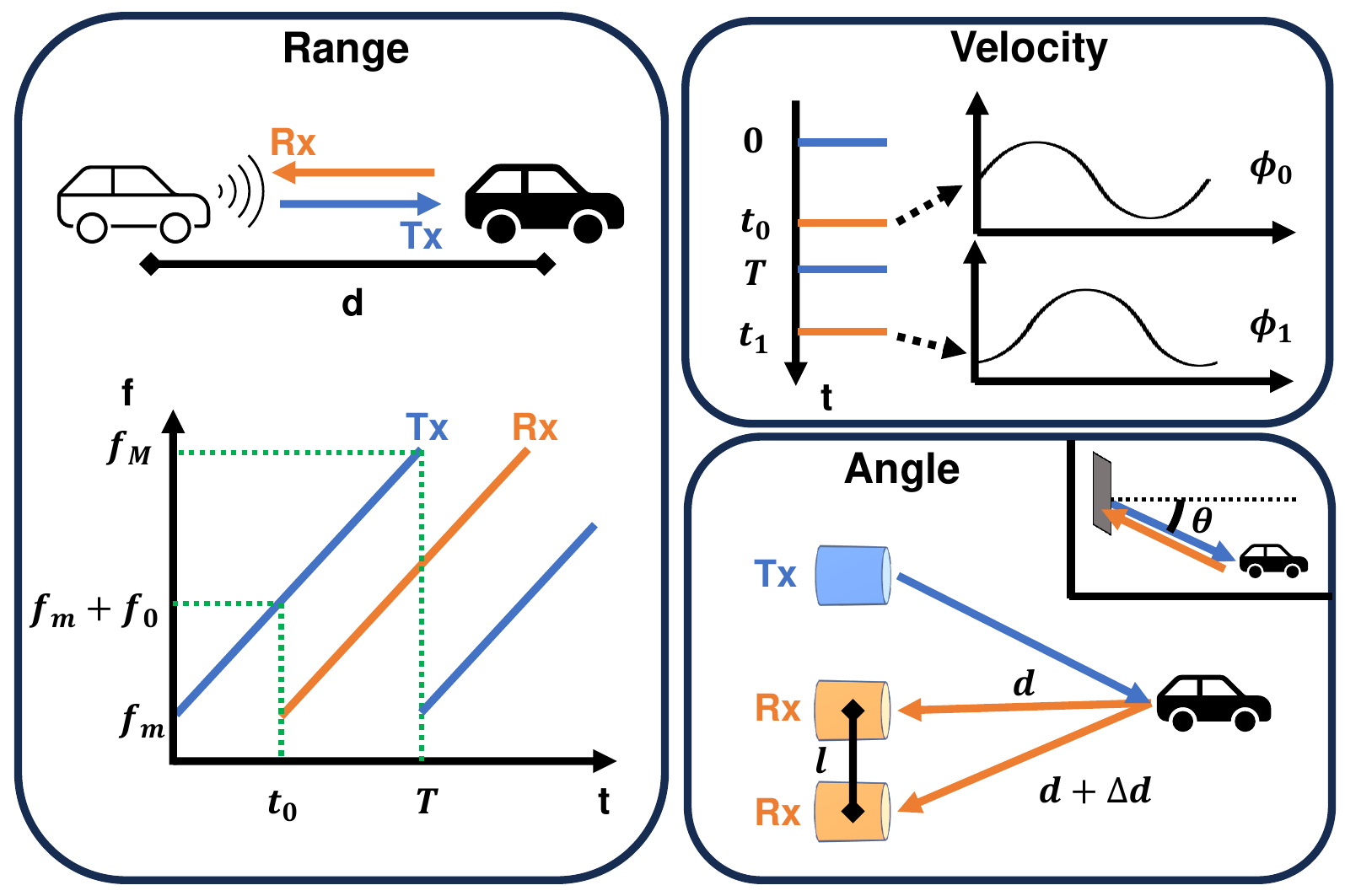}
\caption{Utilizing the FMCW radar, measurements are derived from the chirp signal and the phase of the IF signal. The range measurements are ascertained based on the properties of the received chirp. Both velocity and angle measurements are determined by leveraging the characteristics of the phase difference.}
\label{fig:chirp}
\end{figure}

\subsubsection{Radar Cross Section}
When a mmWave radar sends out an electromagnetic pulse, the waves encounter objects in the environment and reflect back. The \ac{RCS}, which depends on the material, size, and structure of an object, determines the intensity at which each object reflects the electromagnetic pulse. Precisely, RCS is the cross-sectional area of the hypothetical sphere that is generated to measure the reflectivity of the target object. Objects such as vehicles and thick concrete walls will have higher RCS than pedestrians or small obstacles in buildings. The so-called ``radar intensity'' of a reflection from an object is proportional to the latent power, emitted from a radar and reflected off a target in the environment, multiplied by the RCS of the specific target \cite{richards2010principles}. As reported in the literature, intensity served as a valuable measurement for identifying semantic information about objects or in navigation as strong returns are more likely to correlate with recognizable landmarks. 

\subsubsection{\ac{FMCW} Radar}
\label{sec:fmcw_calculations}
Intensity is also determined in part by the antenna used by the radar, and the power and shape of the electromagnetic pulse sent from the radar, along with the antenna's sensitivity to returns from objects.
Antenna can be classified into two types: transmit antenna (TX) and receive antenna (RX). In \ac{FMCW} radar applications, TX antenna are responsible for sending out an RF pulse with linearly increasing frequency, or chirp. These chirps reflect off the environment and the returned signals are captured by the RX antenna. After the RX antenna receives a chirp, the signal is amplified and mixed, or subtracted, from the corresponding TX chirp to create an \ac{IF}. Since both signals have the same slope, the \ac{IF} is a sine wave with a constant frequency.

While the TX and RX antenna can sometimes be the same physical antenna element, most \ac{SoC} sensors separate their TX and RX antenna physically, so as to contain multiple TX and RX antenna. Various chirp parameters including the bandwidth, or change between starting and ending frequency, chirp slope, and the inter-chirp duration can lead to varying performance specifications altering detectable range and velocity resolutions, minima, and maxima. 

Range measurements are derived from the time difference between the transmission and reception of a signal. Using the speed of light $c$ and the first arrival time $t_0$, the range $d$ can be determined by the equation
\begin{equation}
\centering
d = \frac{c}{2} t_0 .
\label{eq:range_meas}
\end{equation}
The left portion of \figref{fig:chirp} shows the region where an \ac{IF} signal is generated using green dashed lines. The difference between the transmitted and received signal is proportional to a constant frequency ($f_0$) sine wave, $sin(2\pi f_0t + \phi_0)$ spanning from $t_0$ to $T$.

The frequency $f_0$ is defined as a function of the distance to the target, the duration of the transmitted chirp $T$, and the bandwidth of the transmitted signal $B = F_M-f_m$. The bandwidth and transmit time are related to the slope of the chirp $S=B/T$
\begin{equation}
    f_0 = \frac{2Bd}{Tc} = \frac{2Sd}{c}.
\end{equation}
This equation can be inverted to solve for the distance as a function of the intermediate frequency \cite{melvin2013principles,iovescu2017fundamentals}.

The phase of the \ac{IF} signal $\phi_0$ can be expressed as an equation,
\begin{equation}
\phi_0 = 2 \pi f_m t_0 = \frac{4\pi d}{\lambda}.
\label{eq:phase_meas}
\end{equation}
While both the base frequency $f_0$ and phase $\phi_0$ are functions of distance $d$, the phase is only valid for sufficiently small distance values and is subject to angle-wrapping. Thus this is typically not used for range estimation but to measure small changes $\Delta d$ where the phase responds linearly in velocity estimation \cite{iovescu2017fundamentals}.

For velocity estimation, at least two sequential chirps are employed as shown in the top right of \figref{fig:chirp}. As the distance in time between each chirp in the sequence is small, on the order of 40 microseconds, the range measurement from both samples and thus the relative \ac{IF} $f_i$ and $f_{i+1}$ are nearly identical.  However, the \ac{IF} signals will possess distinct phases. This phase disparity $\Delta\phi$ corresponds to a motion of the object. Estimated velocity $v$ is determined from the phase difference
\begin{equation}
\Delta\phi = \frac{4\pi \Delta d_{obj}}{\lambda} =  \frac{4\pi v T}{\lambda},
\label{eq:phasediff_meas}
\end{equation}
simplified to
\begin{equation}
v = \frac{\lambda \Delta\phi}{4\pi T}.
\label{eq:vel_meas}
\end{equation}
As this is a function of phase, we note the maximum detectable velocity is unambiguous for $|\Delta\phi|<\pi$. Thus $v_{max} = \lambda/(4T)$ is a function of the wavelength of the signal and the time between chirps \cite{iovescu2017fundamentals}. 

A similar phase difference calculation can be employed for angle estimation \cite{iovescu2017fundamentals}. Given multiple receiver units separated by an interval $l$, the distance disparity $\Delta d$ emerges in reflections. The angle of arrival $\theta$ can be derived from the modification of \eqref{eq:phase_meas} with the geometric relation $\Delta d = l \sin{\theta}$ in \figref{fig:chirp}:
\begin{equation}
\centering
\theta = \arcsin{\frac{\lambda \Delta \phi}{2 \pi l}}.
\label{eq:range_meas}
\end{equation}

While the equations listed throughout this section are conducted in the frequency domain, most signals are captured through digital samples from a high-frequency \ac{ADC} sampling every 100ns or less. The discrete samples are subsequently passed through a series of \ac{FFT} which produce frequency and amplitude graphs which can be analyzed to identify peaks. Range frequencies are identified from an FFT processed on a single chirp and return. Velocity frequencies are calculated from sequences of FFTs calculated across a series of chirp and returns. Angle-of-arrival frequencies are processed by an FFT along TX-RX pairs for given chirps and returns or at a specific antenna angle depending on the radar type.

Frequency modulation acts as a temporal indicator across a continuous wave allowing for range-based calculations to be applied. Compared to amplitude modulation, the accuracy of a returned wave is only affected by the frequency's bandwidth. As such, frequency modulation is less susceptible to concerns related to signal-to-noise ratios and the specific RCS of detected targets affecting the ability to determine individual returns. As FMCW waves are time-multiplexed with unique dwell times between pulses in a pulse train, they act as coherent signals. This means individual chirps can be correlated to the specific TX-RX pair and position in the sequence from which a chirp originated. This allows for correlation of signals in both amplitude and phase versus early time-of-flight pulsed radar where only a non-coherent amplitude measurements could be made. Skilled human operators were required to interpret clutter from signals in these systems. 

%

\begin{figure*}[!t]
\centering
\includegraphics[width=\textwidth]{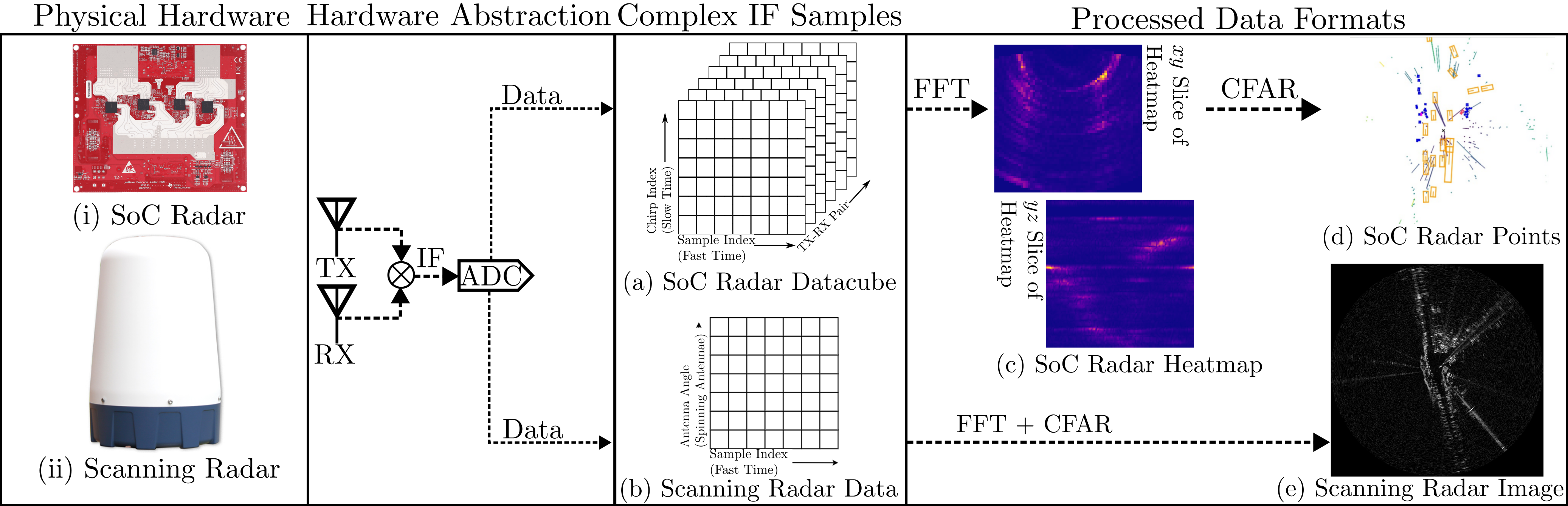}
\caption{The two main types of mmWave radar are (i) \ac{SoC} Radar and (ii) Scanning Radar. 
For both types, transmit (TX) and receive (RX) antenna send and receive signals from the environment. An \ac{IF} is generated by mixing the signals and sampled by an \ac{ADC}. ADC samples can be stored directly. SoC radar stores samples in a datacube (a), where the sample index represents samples along a single chirp from an ADC, the chirp index represent samples along a series of sent chirps, and the TX-RX pair denotes the specific transmit and receive antenna generated a signal. Scanning radar stores samples along a single chirp and by the angle of the antenna at which the chirp was generated (b). These complex IF samples can be furthered processed with SoC radar generating a 3D+1 or 4D Radar Heatmap (c) by applying a \ac{FFT}. Values are stored by range, azimuth, and elevation in spherical coordinates (3D) with each value containing both intensity and radial velocity (+1). A non-maximal suppression \ac{CFAR} filter can be applied to this heatmap to generate radar point returns with radial velocity (d). A similar FFT + CFAR processing generates a measure of intensity along each scanned angle presented in a 2D radar image (e).}
\label{fig:radar_types}
\end{figure*}

\subsection{Sensor Types and Data Formats}
\label{sec:antenna}

As mentioned in the previous section, the antenna types of various radars can have a drastic impact on the generated data formats and other properties each radar possesses. We detail in the following section several of the most common radars that are encountered in robotics research. For each type, we address their main strengths and drawbacks. An example of the two main types of radar and their data types are shown in \figref{fig:radar_types}.

\subsubsection{Imaging Radar}
Imaging radar, or sometimes designated scanning radar, is composed of a physically rotating radar sensor. Imaging radar provide a highly accurate polar image, as shown in \figref{fig:radar_types} (e), of their surroundings by measuring target ranges around in a series of angles. These target ranges are developed by calculating an \ac{FFT} along the \ac{ADC} samples of a given angle and determining the relative peaks in that transform using CFAR filtering. Typically, these radars have long ranges of 100m or more, but without providing calculated Doppler velocity of the returned targets. 

As with lidar, if the sensor is moving along a trajectory, measurements of each angle will be shifted by the relative velocity, leading to a mismatch between the first and last measurements in a single scan. This is more pronounced given that sensors such as the NavTech CIR 304-H have a 4Hz rotation rate. This can either be compensated for accurate image acquisition such as in \cite{burnett2021doppler} or the precise measurement timestamp can be applied for each angle. 

Unlike lidar, however, imaging radar do not provide elevation values of returned targets. This restricts the data to a 2D plane around the scan even if reflections may be returned from various elevations in the antenna waves path. In addition, scanning radar do not typically contain velocity information, as for each antenna angle only a single pulse is sent and received instead of the multiple required to calculate velocity.


\subsubsection{System-on-Chip Radar}
\ac{SoC} radar, combine processing units into a limited number of chips mounted directly patch antennas, or antenna placed directly into a printed circuit board. SoC radar are often lighter weight, and have lower power consumption needs than imaging radar due to their integrated nature. 
The accuracy and resolutions of SoC radar are dependent on the specific antenna array employed, and the manufacturer-specific processing used to correlate measurements across antenna. Antenna patterns have drastically different resolutions along each dimension depending on the geometric differences between antenna in their layout. Equation 5 shows how the angle of arrival is dependent on the length between antenna and when constraining antennas to a small area this distance is often minimized leading to lower resolutions by default. For instance, the TI MMWCAS-RF-EVM, shown in \figref{fig:radar_types} (i), has 4 vertically offset RX antenna admitting a $22.5^\circ$ elevation resolution.

Once these measurements are correlated into intermediate frequencies, an \ac{ADC} samples along individual chirps, along series of multiple chirps, and for each TX-RX pair. The resulting complex IF samples are stored in a data-cube along the sample index, chirp index, and by TX-RX pair as shown in \figref{fig:radar_types} (a). The sample index is a representation of the time along a single chirp of a specific ADC measurement. The chirp index represents the specific wave produced out of a series of waves. An FFT can be run across each dimension of the data-cube to determine the frequencies of the values stored within and processing from \secref{sec:fmcw_calculations} can be applied to generate a heatmap of intensities and radial velocity as shown in \figref{fig:radar_types} (c). This data is typically stored by azimuth, elevation, and range in spherical coordinates. Because radial velocity is thought of as a dimension these heatmaps are sometimes referred to as 3D+1 or 4D radar heatmaps. Systems with limited or no vertical resolution, from having few or no vertically offset antenna, are sometimes referred to as 2D array radar systems and have 2D+1 heatmaps. 

Most SoC systems provide an additional layer of processing beyond the FFT calculation applying a \ac{CFAR} filter with non-maximal suppression to determine individual returned targets, rather than radar images, as shown in \figref{fig:radar_types} (d). These sparse targets are comparable to the point-clouds generated by lidar, but contain an additional velocity dimension. Popular sensors include the TI-AWR1843 2D array radar system, which can also produce CFAR point targets, the SmartMicro UMR-11 and Continental Radar series which produce 4D CFAR filtered targets, or the TI MMWCAS-RF-EVM a cascaded system which produces 4D heatmaps using multiple synced chips acting as a single transceiver.
\begin{figure}[!t]
    \centering
    \includegraphics[width=\linewidth]{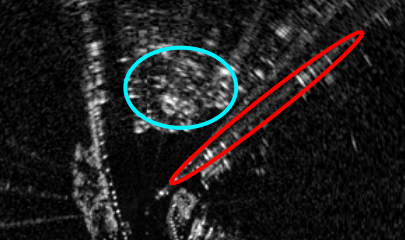}
    \caption{Radar presents several types of noise that are unique compared to other sensors. Speckle noise returns are the most common, with ambiguous clutter circled in cyan. Multipath reflections develop where returns bounce off nearby walls or the ground before hitting the antenna, generating reflections of true targets. A series of repeated returns is circled in red. The original image is sampled from the Mulran dataset \cite{kim2020mulran}.}
    \label{fig:noise}
\end{figure}

\subsection{Challenges to Radar Applications}
\label{sec:radar-challenges}
Like all sensors, radar has many distinct challenges that must be accounted for when developing new solutions.  
Radar has many unique noise characteristics from spurious returns throughout the sensor range, to more complicated speckle noise, and multi-path reflections creating repeated returned signals from a single object. 
An example of several of these characteristic phenomena are shown in \figref{fig:noise}. 

While cluster-based filtering algorithms such as \ac{DBSCAN} \cite{ester1996density}, \ac{OPTICS} \cite{ankerst1999optics}, and the \ac{CFAR} \cite{nitzberg1972constant} help alleviate speckle noise and spurious reflections, they also require precise tuning to individual hardware and antenna patterns.   
Developing algorithms robust to the specific noise characteristics of radar is key to robotics applications. 
We briefly detail common types of noise encountered in radar systems and expose later on in this survey a host of novel methods designed to mitigate its effects.

\subsubsection{Speckle Noise}
After a radar sends out an electromagnetic pulse it receives the reflected energy from objects in the environment. 
As the pulse interacts with environmental objects the radar wave is scattered and the waveform can constructively and destructively interact with itself, either creating spurious returns or negating true returns that the antenna receives regardless of the authenticity of the return. 
Without a way to denote which returns are true or not, the sensor generates a distributions-form of scattered points denoted speckle noise. 
Uncertainty estimation around returned reflections, potentially across multiple scans, is required for precise landmark identification for pose estimation and feature matching.  



\subsubsection{Multipath}
In addition to speckle noise, another type of spurious return called multipath is derived from detection paths differing for one object. 
Consider the case where there is one landmark in front of an autonomous agent. 
Transmitted rays can reach the landmark directly. However, other rays reach the antenna after being reflected from the ground or after hitting a wall. 
From the radar's perspective, the received information of the landmark suggests it exists under the road or behind the wall. 
These can also be designated as so-called ghost objects, or static outliers. 
Removing these outliers is required for reliable point cloud mapping or localization. Multipath noise is present in other time-of-flight sensors and algorithms to address errors, such as \citet{kadambi2013coded}, may have some future applications in radar processing.

\subsubsection{Motion Induced Distortion}
Motion distortion is an inherent characteristic of scanning sensors including lidar and radar. 
Since the imaging radar generates every single polar image from one rotation, the actual vehicle poses at the first ray and the last ray are different if the vehicle is in motion. 
The gap between the first and last ray is nonnegligible when the sensor has a low frequency and/or if the robot moves at high speed.
Frequently utilized imaging radar, such as the Navtech CIR 304-H have a 4Hz frequency, which is challenging for raw frame registration.


As addressed earlier in the introduction, other sensors can compensate for the drawbacks of radar, mentioned throughout this section, providing correlating measurements at various ranges and through various conditions. \tabref{tab:sensor_basics} provides a detailed look at other exterosceptive sensors that might cover the different drawbacks that radar presents.


\subsection{Constant False Alarm Rate Filtering}
\begin{figure}
    \centering    \includegraphics[width=\linewidth]{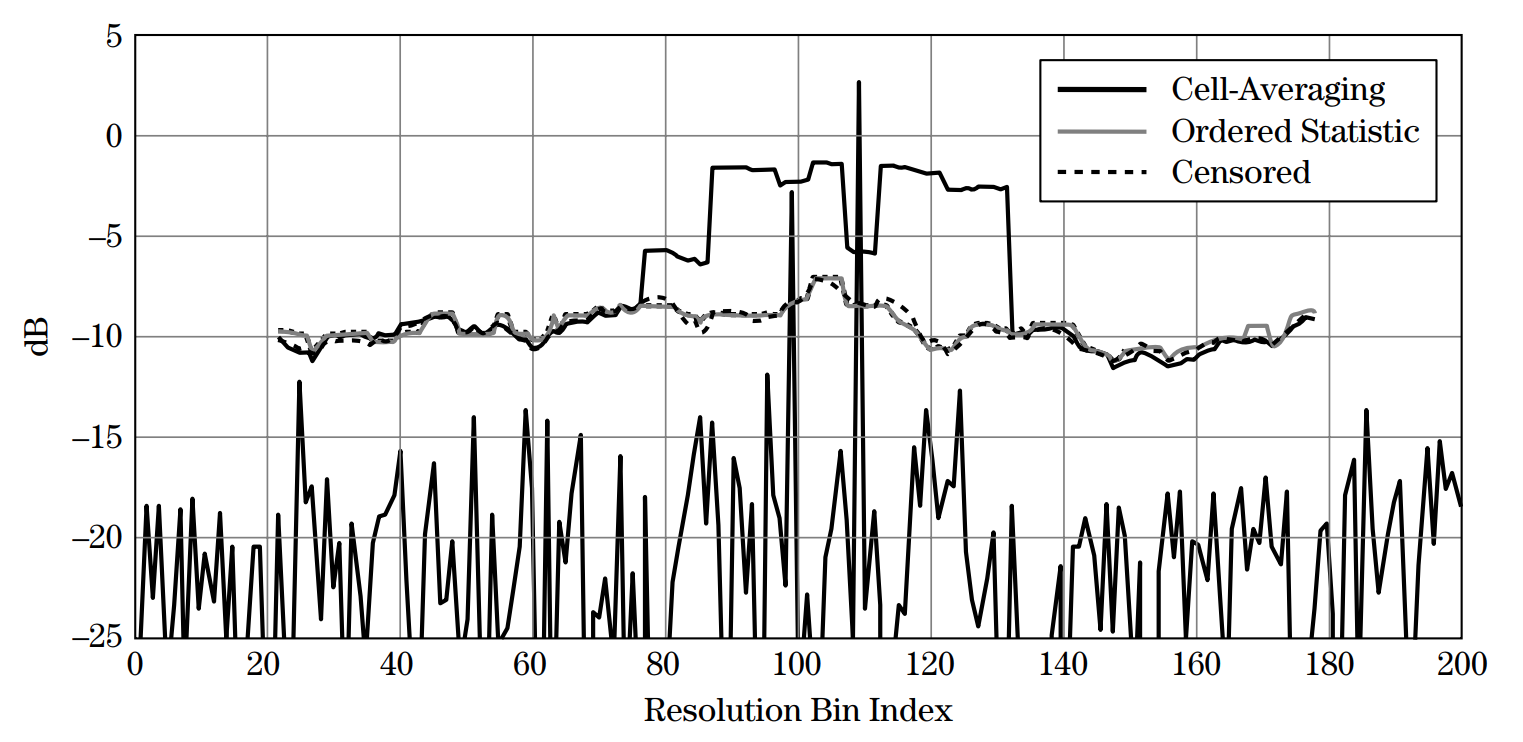}
    \caption{Example of three types of CFAR filtering, Cell-averaging (CA) \cite{finn1968adaptive}, Ordered Statistic (OS) \cite{rohling1983radar,rohling1985new}, and Censored CS(2) \cite{rickard1977adaptive,ritcey1986performance}, applied to a two target detection problem. Figure is adapted from \cite{melvin2013principles}. 
    }
    \label{fig:cfar}
\end{figure}
Radar is subject to a variety of noise, especially when deployed in arbitrary environments. Thermal noise, electronic imperfections, and varying target RCS cause signals to vary across the frequency domain. The highest peaks in the frequency spectrum are thus a combination of true targets and false alarms. The distribution of false alarms is subject to temporal changes with unknown parameters, meaning a static false alarm filter such as threshold will admit false alarms. Researchers have developed techniques to online estimate the distribution of false alarms with these challenges in mind.

One widely adopted estimator is the constant false alarm rate (CFAR) filter, which attempts to maintain a given probability of false alarms in the presence of non-uniform dynamic interference. The first of step is to divide a signal, such as the frequency domain plots generated after applying an FFT to the ADC samples from the radar, into discrete bins called cells. These cells can then be compared with a sliding window across the frequency bin domain. At the center of that window are the set of \ac{CUT}. The magnitude of the CUT is compared to the magnitude of leading and lagging training cells. Optionally, the training cells and CUT may be separated by guard cells, whose magnitudes may be influenced locally by the CUT. A CUT cell set can be accepted or rejected based on whether it contains a magnitude over a threshold developed based on the relative magnitude of training cell sets, excluding the guard cells. Figure \ref{fig:cfar} shows an example of three types of CFAR with cell-averaging CFAR \cite{finn1968adaptive} admitting one return, and ordered statistic \cite{rohling1983radar,rohling1985new} and censored CFAR \cite{rickard1977adaptive,ritcey1986performance} admitting two returns. In all cases the threshold for admitting returns varies across the frequency domain. In this way local noise is accounted for while still admitting high-power returns. 

%% file: motion_estimation.tex
A core problem in robotics is to localize a robot to its environment, and motion estimation is considered the first step in this process. Integrating vehicle motion generates odometry, the change in pose of a system, and obtained odometry is later utilized for organizing robust mapping and localization. In this paper, we first introduce radar-based motion estimation in this section and explain the relocalization and mapping problem in \secref{sec:localization}. Since radar data can be extracted as point-clouds, images, and heatmaps, there are diverse approaches for computing vehicle motion. We divided this section into various methodologies, often aligned with specific data types as a result. 


\subsection{Feature-Based Methods}
Early radar motion estimation methods focused on filtering radar data into traditional SLAM features and applying a unique form of data association between frames to estimate position and generate a map. \citet{adams2012robotic} wrote a book on radar navigation, focusing on probabilistic SLAM and radar filtering techniques. They presented robust estimation techniques such as random finite set approaches to SLAM \cite{mullane2008random,mullane2009random,mullane2010rao,mullane2011random}, prediction techniques \cite{jose2010predicting}, and evidence-based mapping algorithms \cite{mullane2006evidential,mullane2007including}. These methods were designed to be robust to the significant noise present in radar scans. 

As an alternative to probabilistic models, \citet{schuster2016landmark} developed an early GraphSLAM~\cite{thrun2006graph} localization algorithm derived on grid and feature based maps from radar data. 
Landmarks were extracted using \ac{BASD} algorithm. BASD descriptors extend annular statistic descriptors where clusters of pixels, in this case from radar images, are described by statistics in a set of n-rings. 
These statistics are then compared to one another to reduce data storage to binary descriptors. Outlier rejection is accomplished by applying RANSAC \cite{fischler1981random} optimization based on the estimated location against the generated map, and a graph-based optimization improved localization results offline. 



\citet{rapp2016fscd} present a deeper look at the design of a Fast Scatter Center Detector (FSCD), which detects peaks via comparison with nearby grid cells, and pair this detector with BASD on grid-based maps. Through this analysis, \cite{rapp2016fscd} notes that the highest-power centers of radar speckle clusters are consistent landmark descriptors over time, whereas direct grid-map representations of radar data often oversaturate due to the presence of radar speckle noise. Although the use of BASD is quite appealing due to their simplicity to implement, they require precise tuning based on the specific radar's noise characteristics and have trouble generalizing, as noise can propagate across the radar image ``filling" radar grid cells even in otherwise unoccupied space.

Counter to their early work using a grid-based map solutions, \citet{schuster2016robust} present a density-based stream clustering (DBSC) algorithm on radar point clouds to design a particle-filter-based Cluster-SLAM algorithm with time decay to account for dynamic obstacles in long-term localization algorithms. DBSC suffers from similar problems to BASD in terms of outlier rejection, although the authors claim that the sensor decay model helps address clusters created by multipath reflections, which do not maintain consistency throughout a motion trajectory. Fine-tuning may be required to create robust maps and localization. 

Similarly, \citet{rapp2017probabilistic} create a probabilistic estimation framework based on radar clusters to estimate the ego-motion of a vehicle. A Gaussian Mixture Model based on clusters generated by $k$-means, DBSCAN \cite{ester1996density} or OPTICS \cite{ankerst1999optics} is used to maintain spatial estimations, while a single Gaussian model is used to estimate relative Doppler-velocities, maintaining consistency and eliminating noise and dynamics often present in radar scans. This method struggles to estimate off-axis yaw rotations, but the authors note that this might be rectified via additional sensors. 

These early works all focused on using traditional radar processing to generate features that could reliably be applied to grid-map-based localization systems originally designed for 2D laser scanners. They also used the same 2D SoC radars designed for automotive purposes. 


As more radar systems became commercially available, more researchers began looking at radar-based motion estimation. With Navtech's early commercially available automotive scanning radar, \citet{cen2018precise} designed a landmark extraction method that combines attenuated low-frequency and high-frequency data and uses the returned signal strength to identify and remove multipath reflections. These landmarks generated from 1D raw signals are then collated into 2D Cartesian space and a greedy scan-matching algorithm based on pairwise compatibilities 
to maintain geometric consistency aligns multiple clouds. Singular value decomposition is then used to estimate the relative motion between the two scans. Later in \cite{cen2019radar} the unary candidate scan matching algorithm had its computation time reduced by recasting the solution as a global optimization problem across all potential candidate pairs. Broadly, Cen and Newman's works leveraged the long range of radar to design  radar images and localized by aligning persistent geometric features in the radar image space. More recent works such as \cite{lim2023orora} have attempted to improve the resistance to outliers of candidate matching methods. \citet{lim2023orora} used a novel decoupling of rotation and translation estimations. Rotations were calculated using a graduated non-convexity rotation estimator, and they project the resulting scans onto a manifold to conduct anisotropic component-wise translation estimation. These combined methods have better outlier rejection and account for the uncertainty in azimuth information from radar more effectively than standard RANSAC filtering methods. 


Similarly to how the spinning lidar had motion compensation applied, researchers considered this option for scanning radar as well. 
\citet{burnett2021doppler} applied velocity-based motion compensation and Doppler compensation to keypoints from \cite{cen2018precise} and demonstrated how both motion and Doppler compensation can improve estimates for vehicle translation and rotation using radar-based techniques. Both compensations were calculated from an external motion estimator. Overall, motion compensation proved to have a more significant effect than Doppler compensation, at the speeds tested. 

Unlike \cite{burnett2021doppler} who used ego-velocity estimated compensation, \citet{gao2022dc} used radar-estimated Doppler compensation from SoC radar systems for a Doppler-compensated estimation on the features of \cite{cen2019radar}. Features were matched using a novel radar Jacobian transformation that accounts for the transformation of noise from polar to Cartesian spaces to calculate an accurate normal distance transformation with uncertainty estimation. 


Outside of velocity-based filters, some recent works including \citet{luhr2014radar}, \citet{alhashimi2021bfar}, \citet{adolfsson2021cfear}, and \citet{aldera2022goes}, attempt to filter radar features with alternatives to CFAR \cite{nitzberg1972constant} and used RANSAC \cite{fischler1981random} optimized for motion estimation. \citet{luhr2014radar} used binary integration to identify consistent radar returns and speckle noise subtraction, from areas of low target probability, to reduce overall noise in scanning radar image sequences. The denoised images had improved CFAR point detection, with few false alarms, but notably few missed detections. \citet{alhashimi2021bfar} created a bounded-false alarm rate by applying an affine transform to optimally estimate and update probability of false alarm in dynamic environments. \citet{adolfsson2021cfear} use $k$ high intensity points per azimuth as a conservative filter alternative to CFAR \cite{nitzberg1972constant}, and then approximate local surface normals to further reject spurious radar returns. This generates filtered surface points, which can reliably be scan-matched from frame to frame. \citet{aldera2022goes} noted that changes in vehicles positions following a standard kinetic model must have smooth curvature, and created a RANSAC \cite{fischler1981random} filter to reject outlier-matches that do not maintain curvature similarity between scans. 

Some localization methods designed for other robotics sensors have direct applications to radar as well. \cite{adolfsson2022coral} tested a method that could be generalized from lidar to radar. \citet{adolfsson2022coral} used intensity peak-filtered points along with a differential entropy calculation to estimate the alignment between radar or lidar scans. Their differential entropy compared the entropy or ``blurriness" of sequential scans and the aligned final scans. Misaligned scans could be identified by the increase in entropy of a combined scan over its subsequent component scans. 

\subsection{Direct Methods}

Just as how visual odometry can be generated through features or directly from data in algorithms such as DSO \cite{engel2017direct}, some radar researchers have implemented direct methods for mmWave odometry estimation. These methods directly compare or use radar measurements instead of filtering measurements into consistent features. Furthermore, like feature-based methods, the first direct methods spawned in automotive researchers collaborating with Daimler AG. 

\citet{kellner2013instantaneous} used the angular position and radial velocity of a radar sensor to calculate the ego-motion of a vehicle. Stationary targets were classified using RANSAC to identify the largest subset of points with the same linearly related radial velocities. Once filtered, a least-squares model is applied to the stationary targets to fit a velocity estimate of the radar and a single-track Ackerman model is used to convert this into ego-motion estimations. \citet{kellner2014instantaneous} further developed their work by estimating consensus on multiple radars. The additional sensors allow for the estimation of heading in addition to velocity, and further optimized estimates using orthogonal distance regression, which accounted for radial velocity and azimuth position error.


The vast majority of radar-related localization so far has focused on 3DoF navigation, tracking the $(x,y)$ Cartesian position and heading $(\theta)$ of a robot. However, some research has embarked on 6DoF localization tracking the elevation $(z)$, roll $(\phi)$, and pitch $(\psi)$ of a system as well. Using more traditional filter-based estimation techniques, \citet{kramer2020ego_vel} designed a batch-optimized sliding-window ego-velocity estimator based on radar and inertial data that performed to levels similar to visual-inertial odometry systems in ideal conditions and outperformed VIO methods in nonideal conditions, including in darkness. \citet{kramer2020mav_fog} then expanded on their earlier method, creating a radar-inertial odometry system that could perform to levels similar to a visual system, but continued to operate even in dense fog.  

\citet{park20213d} similarly estimate 3DoF velocity. 
\citet{park20213d} used two perpendicular chip-based radars to refine ego-velocity measurements.
As in \cite{kellner2013instantaneous}, RANSAC is applied to estimate static radar targets relative to each sensor. This allows for the removal of dynamic returns that could affect velocity estimation. Each perpendicular radar estimates a 2D velocity along their respective plane, one radar estimates $v_y,v_z$ and the other estimates $v_x,v_y$. The $xy$ and $yz$ inlier points are then used to compute a naive 3D linear velocity estimate through a least squares optimization. Then tangential motion refinement is applied by analyzing zero-radial returns to generate a dominant tangential velocity vector onto which the naive estimate is projected. More recently, \citet{michalczyk2022tightly} designed a tightly coupled EKF-based radar-inertial odometry system. They used stochastic cloning to spatially align scans and extended \cite{cen2018precise} to 4DoF radar measurement correspondence in 3D. Simultaneously, researchers were refining 3DoF motion estimation on scanning radar. \citet{park2020pharao} applied a Fourier-Mellin transform to radar scans, generating decoupled rotation and translation estimates from log-polar and Cartesian radar images, respectively. A course-to-fine phase correlation between Cartesian images further refines the translation estimate.

Several other direct methods employed statistical estimations for odometry. \citet{kung2021normal} used a normal distance-transform metric to track scan alignments to submaps based on a Gaussian mixture model. This method generalized across both scanning and SoC radar systems, and combined with an intensity weighting created state-of-the-art translation and rotation estimates over multiple datasets. \citet{isele2021seraloc} also applied an NDT to do radar based localization, augmented with wheel odometry, with a semantic NDT to compute loop closures aligning labelled radar data. \citet{haggag2022credible} proposed a similar probabilistic approach to estimate the ego-motion using automotive radar. By dealing with the point registration problem with the Gaussian mixture model-based probabilistic algorithm, they could avoid the precise point-to-point matching process while also estimating the credibility of their method. 


\subsection{Learning-Based Estimations} 
In addition to the hand engineered feature-based localization techniques, direct methods, and statistical methods, several works have attempted to localize robotic systems leveraging the unique properties of radar with deep learning. Neural networks can learn consistent features or changes between radar images that are useful for odometry and mapping. These networks have the capability to approximate functions that account for consistent patterns across a variety of data in the presence of non-Gaussian noise. 

Just as with traditional methods, several learning methods were dedicated to identifying robust features for later scan matching to generate odometry. \citet{aldera2019fast} used weakly-supervised learning to temporally relate scans used to filter robust and coherent radar samples for odometry. \citet{barnes2019masking} designed a convolutional neural network to extract a mask capable of eliminating unique radar noise challenges, such as speckle noise and multipath reflections. This mask applied to the Cartesian radar scan results in unique feature maps that are then compared across a set of sample rotations to predict a pose under a softargmax, with pose uncertainty. The only supervisory signal to the system was a ground-truth pose. While \citet{barnes2019masking} computed outstanding performance for odometry estimation, they have high time complexity. Thus, \citet{weston2022fast} replaced the brute force rotation calculation with an FFT-based method. Overall accuracy was lower than \cite{barnes2019masking}, but the computation time decreased dramatically.


Some learning methods began augmenting radar odometry with inertial measurements, just as vision and lidar odometry systems adopted visual-inertial and lidar-inertial methods. \citet{almalioglu2020milli} proposed radar-inertial odometry method using single-chip radar. A radar point association model extracted the landmarks that have enough point-overlap subsets. An inertial module estimated the approximate motion of the body frame in parallel. To conduct UKF-based pose estimation, scan matching was performed with the given point association and inertial motion model. They matched the scans using an NDT-based method, using their RNN-based motion model. \citet{lu2020milliego} fused together motion features from inertial methods using IONET \cite{chen2018ionet} and potentially other deep odometry methods such as DeepVO \cite{wang2017deepvo} together with a mmWave subnetwork that uses a CNN feature extractor designed to track useful mmWave features in a manner similar to optical flow. The motion features are then merged via mixed attention fusion into a recurrent neural network to track 6DoF odometry across trajectories. \citet{lu2020milliego} note that this method works well under low-speed conditions of handheld rigs and wheeled motion robotics platforms, and indoors. They contemplate, however, whether their results will generalize outside of strict indoor environments or the faster motions of conventional MAV platforms.

Rather than training more directly on features, some odometry methods began supervising training directly on odometry or odometry-error metrics. More recently, \citet{barnes2020under} developed a convolutional neural network that learned salient features and their descriptors for odometry based solely on localization error metrics from ground-truth supervision. 
Similarly, \citet{burnett2021radar} used an exactly-sparse Gaussian variational inference model \cite{barfoot2020exactly} to jointly-estimate features, localization data using a data-likelihood minimization technique. Burnett's feature generation technique was adapted from \citet{barnes2020under} with an additional mask on ``empty" feature regions and was unsupervised. \citet{ding2022self} defined three losses for self-supervised radar scene flow estimation, RaFlow. The combination of radial displacements, neighbor correspondence, and spatial smoothness loss each contributed to overall flow estimation.


In another method, \citet{huang2021cross} used lidar-based supervision in training only. \cite{huang2021cross} found consistency between LiDAR and radar using cross-modal contrastive learning of representations and the proposed radar-based navigation. Their results show reliable odometry estimation when compared against existing conditional generative adversarial networks.

Feature generation and odometry methods from applied deep learning techniques provide a promising alternative to traditional filtering methods, given their lower reliance on hyperparameter tuning. But methods may still struggle to adapt to unseen environments, i.e. from city to country driving, or differing sensor modalities. As in other learned domains, care in cross-validating methods with a reliance on deep learning will be required.




%% file: relocalization_and_mapping.tex
After determining frame-to-frame motion estimates, it is necessary to relocalize and generate maps for long-term navigation. This can take the form of the so-called \textit{kidnapped robot problem}, where the robot must align itself to an existing map after being brought to a new location. It can also represent loop closures in SLAM where trajectory estimates are improved through repeated visits to the same location. In both cases, robots rely on consistent maps which may be used for current and future estimates along with motion plans. 

\subsection{Localization}
Many relocalization algorithms relied on scanning radar, taking advantage of the long ranges provided to identify consistent geometry. \citet{hong2020radarslam} conducted parallel processing for radar-based loop closure and motion tracking. Since the bag-of-words method is hard to apply in sparse radar images, they converted the data into a point cloud to conduct a loop closure based on \ac{M2DP} descriptors. For motion tracking, they first transformed polar images into Cartesian images to find feature points. The transformation matrix for motion estimation was then obtained from the maximum inlier feature point set. 
\citet{jang2023raplace} introduced a radar place recognition technique employing the Radon transform. They established a mutable threshold through an auto-correlation procedure, exhibiting superior robustness to both rotational and translational variances.
\citet{adolfsson2023tbv} improved a SLAM algorithm by introducing a trust but verify-based loop closure detection system. While generating loop closures with odometry and radar scan description-based methods, a group of local loop closures was kept for verification. This verification optimized for the best overall alignments and effects to odometry estimates providing more accurate long-term navigation. 

Several scanning radar works used machine learning to identify robust relocalization. \citet{suaftescu2020kidnapped} used cylindrical convolutions in a convolutional neural network (CNN) to perform radar place recognition on repeated trajectories. Raw radar scans improved recall over existing image recognition pipelines. \citet{gadd2020look} learned rotation invariant representations of radar images using standard neural nets and compared them with \cite{suaftescu2020kidnapped}. These rotation invariant matrices were then processed into difference matrices to perform backward loop closure detection in SeqSLAM \cite{milford2012seqslam} for relocalization with mirrored search to account for visiting trajectories from opposite directions. \citet{wang2021radarloc} generated a learned relocalization pipeline which filtered radar data using self attention and encoded features using a DenseNet \cite{huang2017densely} architecture processed by a final deep pose regressor. Their method outperformed several existing relocalization techniques in translation and performed at parity for rotation. In a unique SoC mapping method, \citet{cait2022autoplace} filtered single-chip automotive radar using Doppler measurements and embeds the filtered pointclouds using a convolutional NN for spatial encoding. They then applied a single-layer LSTM for temporal encoding to generate and match query candidate pairs of radar cross-sections for relocalization. 

As in the motion estimation papers, some localization manuscripts used external sensors for supervision. \citet{yin2021radar} used lidar based submaps, and a low frequency components of an FFT applied to a birds-eye-view of radar information to jointly train neural networks using a single triplet loss function to generate heterogeneous place recognition that was generalized across data sets along with lidar and radar sensor modalities. Place recognition is a common technique to create loop closures in the SLAM algorithm, or when performing localization on existing maps. \citet{burnett2022radar_lidar} directly compared radar, lidar, and cross-modal mapping and localization techniques. While radar was capable of performing at relative parity to lidar, filtered lidar still outperformed radar-based localization even in moderate weather. This demonstrates the need for potential refinement in filtered-feature-based radar methods being required.

Contrary to previous methods, based on existing radar or lidar maps, \citet{tang2020rsl} designed a system to align scanning radar with overhead satellite images. Their method first estimates the rotation between radar and satellite images via a rotation selector network, and then generates a synthetic radar image that can be used to calculate the offset from the overhead image.

\subsection{Mapping}
Mapping with radar can be difficult due to the spurious returns present. Some of the first works to address mapping, such as \cite{mullane2006evidential,mullane2007including}, attempted to form evidence-based maps to reduce the impact of noise in standard Bayesian formulations. \citet{mullane2006evidential} demonstrate how existing target detection models contain sufficient information to inform evidence-based occupancy grid-map estimates rather than through a priori Bayesian estimates on mmWave radar. \citet{mullane2007including} integrated detection probability estimates from mmWave power-spectral data into map building. They demonstrated both a direct target presence probability algorithm based on signal-to-noise ratio estimates for single shot probabilistic map estimates, and a recursive likelihood estimate based on the output of ordered-statistics CFAR point targets for multi-view probabilistic map estimates. 

Other early work in radar mapping used radar-lidar fusion. \citet{fritsche2016radar} fused the sensors based on estimated ranges to determine which sensor to trust in a given case. This algorithm is moderately robust to localized fog, but would likely struggle in larger environments given that at longer ranges it inherently trusts radar and requires precise knowledge of sensor characteristics to determine when to use the line scanning radar or lidar. \citet{nuss2018random} designed a state estimation filter to address dynamic obstacles in grid maps called a probability-hypothesis-density, multi-instance Bernoulli filter. This filter casts grid cells as a finite stochastic set, and fused radar and lidar data dynamically. They test this filter in a tracking application on a slow-moving vehicle with several dynamic obstacles in the scene. Each of these works attempted to fuse radar data with lidar data, a common localization sensor with relatively high range and accuracy. Combining lidar with a sensor which often has equally high range but which does not fail in the presence of visual degradation would allow for consistent localization across a broader range of conditions, leveraging the best sensor to use under varying conditions. Instead of fusing radar and lidar, \citet{xu2022learned} focused on developing a regression model based on LiDAR supervision, with the objective of accurately estimating radar depth information to facilitate 3D indoor mapping using radar. Their model incorporates robust depth filtering during the training process and extracts valid points.

Researchers have more recently begun developing radar-only mapping methods as well. \citet{cheng2022novel} created a GAN to generate point clouds based on range-Doppler velocity matrices using lidar as a critique signal which outperforms traditional FFT-CFAR filtered pointclouds in new a similarity metric with lidar pointclouds. \citet{kramer2020mav_fog}, in addition to generating odometry, presented a novel sensor model for voxel based mapping of radar data and capable of creating sparse maps as it traversed based on radar data even through visual occlusions. The sensor model leveraged the log-odds based estimation of occupied vs free cells used in \cite{hornung2013octomap}, but replaced the ray-cast model. The ray-cast model assumes that the first contact of a sensor is the only relevant point of occupation, but the generalized model accounts for radar's ability to penetrate certain material types by updating voxel probabilities within the sensors field of view, increasing probabilities in cells with radar returns and decreasing probability with missed scans, without assuming information along a ray. 


As with other sensors, some works have also attempted to add semantic information to their maps. \citet{lu2020see} classified segments of interest along the FFT-profile of radar returns to identify walls, windows, and doors as common semantic features of the house. \citet{isele2021seraloc}, previously mentioned in \secref{sec:localization}, used a reduced subset of SemanticKITTI labels \cite{behley2019semantickitti} and used the labels to spatiotemporally filter points for NDT-scan matching for localization and for more robust loop closure identification.

%% file: object_classification.tex
Aside from navigation where state estimation and mapping are foundational techniques, an additional consideration for robotic sensors is whether they can detect the types of objects in their environment. There has also been considerable effort in imbuing radar with this capability, either by fine-tuning methods developed for other sensors (e.g., vision-based techniques), or by developing completely novel techniques for semantic classification. In this section we will describe some of these methods, dividing our review by those that perform their tasks with the radar alone or if they fuse other sensors to accomplish this.

\subsection{Radar Only Semantics}
Radar data has been used to identify and classify objects directly. These classification methods are being broadly categorized by their use of either processed radar targets, or the less processed radar heatmaps, generated by various systems. 

\subsubsection{Radar Target Classification}
Neural networks based on CFAR filtered radar pointclouds attempt to classify the sparsely available points. They often rely on the additional velocity and intensity information available with each point to generate these classifications. For instance, \citet{nordenmark2015radar} designed one of the first machine learning-based solutions for object classification, using a mixture of cluster filtering techniques such as DBSCAN \cite{ester1996density}, and classical machine learning techniques including support vector machines \cite{cortes1995support}, and principal component analysis \cite{wold1987principal}, to identify bicycles and vehicles for automotive purposes using Doppler velocity information. \citet{danzer20192d} used a PointNet-based neural network to classify cars versus clutter in 2D radar point clouds separated based on a patch proposal method. The authors noted that wheel velocity, which was often different from the vehicles' velocity became a key features in separating cars from general radar clutter. 

In a general perception study, \citet{zhao2019mid} trained a neural network on a set of hand tuned DBSCAN \citep{ester1996density} based clusters to identify and track unique users in a smart-space, while limiting privacy concerns present in many visual-based methods. Similar systems could be used by robotics applications to identify dynamic presences in human robotic interactions.
Recently, \citet{zeller2022gaussian} proposed the Gaussian transformer-based semantic segmentation, which is designed to handle radar noise. Their attentive sampling modules capture the spatial relations, addressing some of the limitations of the standard transformer model in identification of various vehicles and static scenery.

\subsubsection{Radar Image Classification}
Radar images, or heatmaps, are appealing to researchers given their similarity to images, especially in classification. Given that the radar data is often a top-down representation of a scene, objects can also be accurately tracked through multiple images if desired. Both \cite{major2019vehicle} and \cite{akita2019object} used recurrent neural networks to track objects identified in earlier convolutional neural networks through scenes,  demonstrating a cross between metric and semantic applications. 

\citet{major2019vehicle} avoids peak detection methods like CFAR and non-maximal suppression in favor of creating direct radar tensors to train a vehicle detection network. The network separates the full 3D range-azimuth-Doppler tensor into multiple bimodal, range-azimuth, range-Doppler, and azimuth-Doppler tensors with intensity summed across the removed dimensions. These ``images'' are then concatenated and the features space is converted from polar to Cartesian coordinates before applying time varying information in a traditional LSTM \cite{hochreiter1997long} block to generate detections and velocity estimation of other vehicles. \citet{akita2019object} showed how an LSTM-RCNN could be used to classify and track vehicles, pedestrians, and cyclists on raw radar range-azimuth images. Additional processing of these images including hand selected feature clouds and cluster maximums reduced accuracy metrics across the neural networks. Rather than separating 3D tensors, \citet{palffy2020cnn} conducted single-shot obstacle detection of pedestrians, cyclists, and vehicles on sub-samples of range-azimuth-Doppler velocity radar images processed by a two stage CNN analyzing spatial and velocity data around identified targets.

\citet{wang2021rodnet} used stereo-object localization fused with radar range-azimuth frequency heatmaps in a cross-modal supervision pipeline to a secondary radar only neural network which produce confidence maps of prediction labels on said heatmaps. These then undergo a localized non-maximum suppression to identify distinct object center peaks. 
\citet{wang2021rodnet} identifies weaknesses in detecting larger objects and detecting multiple objects, provided the radar's limited resolution in some dimensions.


Outside of automotive-centered classification, \cite{lien2016soli} use high frequency dynamic clustering algorithms to track gestures from a singular wide angle beam, leveraging digital beam steering techniques and machine learning to identify said gestures via feature extraction in cluster centers. The authors highlight the use of velocity information as a key breakthrough in identifying relevant signals in otherwise low resolution positional signals from radar spectral images.

\subsection{Radar Fusion}
Fusion of various sensors with radar can leverage the benefits of each sensor while mitigating drawbacks. In semantic applications, cameras and lidar can provide a level of detail that radar sensors are incapable of replicating due to its lower resolution. But radar can assist these sensor in tough environmental conditions, and leverage the velocity measurements of systems to add additional feature identification not captured in static images or point clouds. Notably, many of the methods developed for classification are centered around the identification of vehicles, pedestrians, and cyclists in autonomous driving applications. 

A common pairing for semantic labelling applications are cameras and radar. Cameras provide information about color and higher resolutions than radar, while radar will often have more accurate range estimates, especially at a distance. When fusing cameras and radar, \citet{chang2020spatial} enumerated three potential methods: a \textit{decision level} or \textit{late fusion} method, where predictions from multiple sensors are fused; a \textit{data level fusion} where regions of interest are generated by one sensor and applied to the other; and an \textit{early fusion} method where features from each sensor are transformed into a common frame of reference. 

Some of the earliest radar applications were in self-driving vehicle settings. Many of those were radar-camera fusion applications to identify vehicles in the environment and relied on data level fusion. These methods included camera to radar calibration for obstacle detection and visualization in \citet{sugimoto2004obstacle}. \citet{fang2002depth} fused radar range detections across a series of histogram bins to enhance depth data acquisition for edge-based binocular segmentation algorithms. A similar method, proposed by \citet{bombini2006radar}, suggested using radar points to inform search conditions for vertical edges in camera based images to identify vehicles. These older methods are held back by requirements for hand engineering and region of interest identification that have limited adaptability. 

A renaissance of radar-camera fusion methods began with the advent of neural networks that eliminate much of the need for heavy hand engineering and achieves higher accuracies. Learned methods often applied late fusion of generated features which could be concatenated or processed in further neural layers. These methods are often rapidly iterated upon as new architectures, generated for visual systems, continue to improve upon accuracy and precision. 

Some early algorithms including \cite{chadwick2019distant,meyer2019deep} developed end-to-end fusion networks around radar. \citet{chadwick2019distant} showed that augmenting multi-focal length camera based detections of distant vehicles with radar target data created higher average precision recall than visual only methods across a range of object sizes. To accomplish this \cite{chadwick2019distant} projected radar points into image space trained in a separate branch of a neural net architecture and concatenating those radar features into a existing visual ResNet18 \cite{he2016deep} architecture. \citet{meyer2019deep}, in addition to releasing a small dataset containing radar and lidar point clouds along with camera images, explored using convolutional neural networks to identify 3D bounding boxes of cars through radar-camera-based fusion. They showed that radar-based fusion outperformed the average precision and recall compared to lidar-based fusion even when partially- or fully-occluded.

Other authors such as \cite{john2019rvnet,chang2020spatial} relied on popular image segmentation networks and modified them for radar fusion. \citet{john2019rvnet} used radar point clouds projected into the camera's coordinate system with ego-velocity compensation on the Doppler velocity dimensions to create a sparse radar image processed by a neural network and fused with a YOLO \cite{redmon2016you} based image features output by a fusion branch. Their proposed RVNet outperformed the average precision of YOLO in a binary classification setting. It identified obstacles even in adverse conditions, but struggled in multiclass framework, e.g. identifying pedestrians, vehicles, or cyclists. \citet{chang2020spatial} merged radar and image based data at the feature level of a CNN using a newly designed spatial attention fusion method that extracted various spatial features from radar data using multiple convolutional kernel sizes in parallel to identify features in the radar data. The output of this radar spatial attention model was then concatenated with an image detection model and further processed in a fusion network. Interestingly, the fused data is then processed by an existing image detection network, as if the data has been early-fused. \citet{kowol2020yodar} applied a late fusion of a 1D radar segmentation network and a YOLOv3 \citep{redmon2018yolov3} object detection network in an uncertainty aware manner, improving detection of objects at night in the presence of nonuniform lighting conditions. 


A notable exception to late fusion methodologies came from \citet{nobis2019deep}. This work developed a radar-camera fusion method where radar data across several samples are fused into image space as perspective-adjusted vertical bars at the aligned pixel locations. The early fused data is then passed through a neural network architecture that online determines the best fusion weights for each sensor, by passing the data across the length of the network. This method allowed them to outperform state-of-the-art visual identification methods across multiple datasets, with radar data helping in the presence of visual degradation from weather. 

While less common, radar-lidar fusion methods have also been developed, often with tracking objects through scenes as a primary goal. \citet{yang2020radarnet} developed a network that combined the complimentary information between lidar and radar to identify and track the velocities of vehicles, and motorcycles for self-driving applications. They leveraged an early voxel base combination scheme to develop birds eye views of the relevant point clouds for a neural net. \cite{yang2020radarnet} then added late fusion of radar velocities with ego-motion compensation to refine the velocities and tracking of classes using an attention-based fusion method. This work shows the advantages of fusing velocity information into semantic labels. 

In a unique application, \citet{radwan2020multimodal} created an interaction-aware temporal convolutional neural network to track pedestrians and vehicles using radar and lidar based fusion to inform street crossing safety for last-mile delivery robotics deployed in arbitrary environments. The tracks are then late fused with a visual identifier of traffic signals generated from a separate neural network. Both methods individually perform or outperform state-of-the-art methods and lead to high precision and accurate decision-making in safe street crossing models.

%% file: datasets_and_calibration.tex
As radar is a relatively new sensor in robotics, many research labs may not yet have purchased a radar system for research. To assist in the discovery and advancement of radar-related methods, a number of groups have released datasets for radar. We distinguish each in Table \ref{tab:datasets} based on the radar characteristics, additional sensors, and broad environment the radar was deployed in. Furthermore, we provide descriptions for various radar datasets separated into imaging and SoC radar sets. Some researchers have also created mmWave simulators. If a lab already owns a sensor, they may wish to accurately calibrate it against other sensors on a robotic system or sensor rig, so we also detail several recent calibration works which use unique targets and optimization algorithms to better align multiple sensors.  

\input{tab_dataset.tex}

\subsection{Datasets}
\subsubsection{Imaging radar dataset}
Scanning radar datasets for localization include \citet{barnes2020oxford} who released an early scanning radar dataset covering 280km of mixed weather urban driving across trajectories previously covered in the Oxford RobotCar datset \cite{maddern2017oxford}. Later, \citet{burnett2022boreas} created a 385 km dataset in adverse weather including snow and rain for autonomous vehicle navigation. This dataset used a scanning radar, lidar, and mono-camera with post-processed RTX groundtruth. The goal of this dataset was to improve the localization in highly diverse weather conditions and in more diverse environments than the urban center covered in \cite{barnes2020oxford}. 

For relocalization, \citet{kim2020mulran} released a dataset focusing on scanning radar and lidar with 6DoF groundtruth trajectories. Trajectories in this dataset included high structural and temporal diversity to emphasize use of range-based place-recognition algorithms, where the longer measurements afforded to radar helped it outperform lidar in place recognition tasks. 

Lastly, \citet{sheeny2021radiate} released a scanning radar dataset in adverse weather to emphasize object detection, tracking, and scene understanding with 200K 2D annotations on road actors in various driving scenarios. 

\subsubsection{SoC radar dataset}

\citet{caesar2020nuscenes} provided SoC radar dataset for machine learning. Five radars with one LiDAR and six cameras acquire urban autonomous driving data. Massive data quantity and annotations are appropriate for object detection and tracking. Traversal data is separate from semantic data, however, and this makes it hard to validate long-term odometry methods.

\citet{schumann2021radarscenes} published another SoC radar dataset with four radars and one camera. Points from four radars and camera images are labeled for 11 road objects, and each data sequence has a single-place trajectory. Some of the sequences can be utilized for simple odometry; however, it is hard to test loop closure. The absence of IMU and LiDAR can be a disadvantage for practical odometry estimation.

\citet{kramer2021coloradar} designed a chip-based radar dataset for 6DoF localization with multiple entry points that include raw ADC values, 4D heatmaps, and radar point targets from both a cascade and single-chip radar. Each dataset contained lidar SLAM-based ground-truth localization. 

\citet{paek2022k} released a 4D radar dataset for object detection on multiple weather environments. The intention behind this initiative is to encompass a wide spectrum of driving environments, which includes urban and suburban areas, daytime and nighttime scenarios, and adverse weather conditions. The dataset can be utilized for demonstrating the resilience of the radar system in challenging environments.

\citet{palffy2022multi} created a 4D radar, lidar, and stereo-camera dataset with 3D bounding-box annotations of 13 classes of vehicles, pedestrians, and cyclists. They also released baseline tests of a classification network based on PointPillars \cite{lang2019pointpillars} using radar pointclouds.

\subsection{Simulation}
Simulation of \ac{mmWave} radar has been limited as the complex interactions between materials and electro-magnetic waves can be difficult to model. Some recent works have attempted to generalize simulations for robotics applications, however, so users can test or train models with radar, prior to acquiring hardware.

In an early work \citet{dudek2010millimeter} designed a \ac{FMCW} \ac{mmWave} simulation environment using a ray tracing channel simulator based on an Agilent assistive driving sensor that operated in Matlab.  

\citet{li2020research} designed a simulator that accurately represented ground and weather clutter in intelligent driving simulations. They were able to simulate range-Doppler plots for a variety of vehicle based scenarios.

Most recently, \citet{schoffmann2021virtual} designed ViRa a virtual radar which simulates raw radar data generate from multi-antenna arrays measuring both conductive and non-conductive surfaces,  including humans and obstacles, in simulated environments.

\subsection{Calibration}
Calibration between radar and other sensors is necessary to maintain consistent and accurate results in multi-sensor fusion applications. Various methods of target-based and data-driven fusion of radar to camera and lidar have been presented. 

\subsubsection{Artificial Target-based Calibration}

Several calibration methods opted to design calibration targets that were detectable by mmWave radar, lidar, and cameras. \citet{pervsic2019extrinsic} designed a radar-lidar-camera target and a multistep calibration optimization. The target relied on a triangular corner retro-reflector with a unique radar cross-section and a transparent foam board with a checkerboard pattern for lidar and camera calibration. This board is first used for a projection error optimization and then the radar cross-section difference resulting from radars varying power projection from 0 elevation is used to refine high uncertainty calibration parameters, namely elevation, pitch and roll. 

\citet{lee2020extrinsic} conducted radar to lidar calibration in a two-stage method, first reprojecting lidar point clouds into the central radar plane, and then using an elevation compensate radar cross-section metric to optimize for pitch, roll, and elevation errors. A second temporal adjustment to lidar using linear interpolation along the proposed trajectories matched target returns along the azimuth of each sensor. 

\subsubsection{Arbitrary Target-based Calibration}

Calibration without specific targets has also been analyzed. Allowing the use of arbitrary targets allows for a broader application of calibration to existing datasets or researchers without access to precise manufacturing capabilities. \citet{scholler2019targetless} used a data-driven approach to combine radar point targets projected onto a camera plane using a coarse-to-fine neural net to determine the rotation between the sensors.  

\citet{wise2021continuous} used radar velocity estimates and camera pose estimates to jointly estimate sensor extrinsics between 3D radar sensors and monocular cameras undergoing continuous time rigid body motions. This method calculated velocity on arbitrary radar returns, without an explicit requirement for radar retro-reflectors.  

\citet{pervsic2021online} created a dynamic object based calibration algorithm. Each sensor would detect and track objects through a scene, and track-to-track association was used to calibrate sensor-to-sensor extrinsics. This graph-based extrinsic calibration was run after an online rotational calibration disturbance was detected. 

\citet{pervsic2021spatiotemporal} used Gaussian processes to jointly track targets between multiple sensors, interpolating across the trajectory to temporally align estimates before applying an on-manifold iterative-closest-point optimization to calibrate sensor extrinsics. 

\subsubsection{Trajectory-based Calibration}
Some researchers chose to ignore target-based calibration entirely, instead optimizing calibration based on odometry or the ego-velocity of particular sensors. 

\citet{wise2022spatiotemporal} adjusted their work from \cite{wise2021continuous} to be target independent. \citet{wise2022spatiotemporal} estimate B-spline control points using a continuous batch-optimization procedure operating on radar ego-velocity and camera odometry estimates.  \citet{cheng2023extrinsic} calibrate 2D automotive radar pairs using a batch calibration between time synchronized radar ego-velocity estimates.

%% file: tab_dataset.tex
\begin{table*}[]
\resizebox{\linewidth}{!}{%
\begin{tabular}{ll|llll|ll}
\hline
 & \textbf{Dataset} & \textbf{Lidar} & \textbf{Cameras} & \textbf{Groundtruth} & \textbf{Semantic Labels} & \textbf{Environment} & \textbf{Inclement Weather} \\ 
\hline\hline
\multicolumn{1}{l|}{\multirow{4}{*}{\textbf{Scanning Radar}}} & Oxford Radar RobotCar \cite{barnes2020oxford} & Yes & Stereo/Mono & GPS/IMU + VO & No & Dense Urban & Rain and Fog \\
\multicolumn{1}{l|}{} & Boreas \cite{burnett2022boreas} & Yes & Mono & GPS/IMU + RTX & No & Sparse Urban & Rain, Snow, and Fog \\
\multicolumn{1}{l|}{} & MulRan \cite{kim2020mulran} & Yes & No & SLAM  & No & Mixed Urban & N/A \\
\multicolumn{1}{l|}{} & RADIATE \cite{sheeny2021radiate} & Yes & Stereo & GPS/IMU & 2D Boxes (8 Categories) & Mixed Urban & Rain and Snow \\ 
\hline
\multicolumn{1}{l|}{\multirow{4}{*}{\textbf{SoC Radar}}} & nuScenes \cite{caesar2020nuscenes} & Yes & Stereo & GPS/IMU & 3D Boxes (23 Categories) & Mixed Urban and Natural  & Rain \\
\multicolumn{1}{l|}{} & RadarScenes \cite{schumann2021radarscenes} & No & Mono & None & Point-Wise (11 Categories) & Mixed Urban Roadways & Rain and Fog \\
\multicolumn{1}{l|}{} & ColoRadar \cite{kramer2021coloradar} & Yes & No & SLAM & No & Varying & N/A \\ 
\multicolumn{1}{l|}{} & K-Radar \cite{paek2022k} & Yes & Stereo & GPS/IMU & 3D Boxes (5 Categories) & Mixed Urban and Natural & Overcast, Fog, Rain, Sleet, Snow \\ 
\multicolumn{1}{l|}{} & View-of-Delft \cite{palffy2022multi} & Yes & Stereo & GPS/IMU & 3D Boxes (13 Categories) & Mixed Urban & N/A \\
\hline
\end{tabular}%
}
\caption{Overview of various radar related datasets, including other sensor types, locations, and encountered weather. \label{tab:datasets}}
\vspace{-7mm}
\end{table*}

%% file: openproblems.tex
In the robotics community, radar applications have been rapidly adopted and iterated upon. However, there are still open research questions. More specifically, \ac{mmWave} radar sensor research has developed most rapidly in the industries for which they were first designed, namely autonomous vehicles. Planar localization and mapping in metric applications, and classification of vehicles, pedestrians, and other common ``traffic obstacles" in semantic applications are rapidly maturing research directions. 

Still, the accuracy of planar radar-based localization systems still lags that of lidar systems (approximately 10 cm error as compared to 5 cm).  This may in part be improved through maturation of radar sensors themselves in terms of higher resolution and higher scanning speeds. However, it may also be possible to make algorithmic improvements through different choices of map representations, features, and back-end state estimators.  Place recognition for robust loop closure detection using radar is another relatively unexplored area.  Visual and lidar odometry and localization benefited greatly from the public KITTI \cite{geiger2013vision} benchmark, which spurred many researchers to work on these problems.  We hope radar odometry and localization benchmarks such as the Boreas dataset \cite{burnett2022boreas} will do the same for radar.

Planar perception alone is often insufficient to create long-term repeatable actions in several domains. While early research in expanding the degrees of freedom for localization systems to 6DoF localization has begun in earnest with early promising results, it has not yet been applied to common UAVs with high relative speeds. Here again improvements in the sensors may be required in terms of field of view, resolution, scanning speeds, as well as physical specifications such as size, mass, and power. Additionally, work generating robust 3D mapping useful for planning algorithms is necessary for radar to become a primary sensor in many robotics cases. It is an open problem to build a highly robust 6DoF radar odometry and localization solution capable of working in real time in a wide variety of environments.

For semantic applications, broadening classification targets beyond just vehicles and pedestrians has begun, with more detailed models including differentiation of vehicle types. Increasing the average precision of systems at distance will likely continue to be a major focus. Adding additional domains to radar semantics is still an open research area. New semantic labels for offices, homes, hospitals, and other non-driving focused applications will enable radar to be used in a wider variety of environments.  \change{Panoptic segmentation of radar scans seems to be a under-explored area, which could be aided greatly by Doppler velocity.} 

While there have been some approaches to offline radar calibration, online techniques have been relatively unexplored.  It would be exciting to see additional work in this area, particularly intrinsic calibration 
to allow radars to be treated more like a commodity 
rather than custom units.

Finally, there are some exciting opportunities to make better use of the full waveform that is available with radar.  This is common in ground-penetrating radar surveying, for example, using classic radar theory.  But perhaps there are some clever ways to combine modern representations such as neural radiance fields with radar waveforms to produce better maps in challenging environments, including behind building walls or in other occluded regions.  Radar has a lot to offer beyond basic pointclouds that we can surely unlock by being creative.

%% file: conclusion.tex
Throughout this survey, we have demonstrated how mmWave radar has advanced both in terms of its hardware capability and how algorithms have adapted to this new class of packaging. We have conducted this analysis in the context of adding radar systems to robotics applications. Radar, notably, is already a vital sensor in automotive safety systems and has great potential at becoming a core exteroceptive sensor in robotics, similar to the camera and lidar. Radar sensors have already significantly decreased in price, with some automotive radars priced similarly to machine-vision cameras. Beyond ease of access, radar provides benefits both through fusion and as a standalone sensor, bypassing particulate matter in visually degraded environments and adding additional information via the Doppler-velocity channels. 

Radar has become a core robotics sensor in applications for operating in visually degraded environments. As the reliability and accuracy of radar sensors increase and their prices decrease, the prevalence of these sensors will continue to grow. This work serves to show that even in the early days of these small form-factor, low-power sensors, exciting robotics applications of mmWave radar are beginning to take shape.